\documentclass[journal]{IEEEtran}

\usepackage[utf8]{inputenc}
\usepackage{float}
\usepackage{graphicx}
\usepackage{booktabs}
\usepackage{cite}
\usepackage[dvipsnames]{xcolor}
\usepackage[normalem]{ulem}
\usepackage{color,soul}
\usepackage{adjustbox}
\usepackage[lofdepth,lotdepth]{subfig}
\usepackage{amsmath}
\usepackage{algorithm}
\usepackage{algpseudocode}
\usepackage{multirow}
\algrenewcommand\algorithmicrequire{\textbf{Input}}
\algrenewcommand\algorithmicensure{\textbf{Output}}

\begin{document}
\title{Multiple-Perspective Clustering of Passive Wi-Fi Sensing Trajectory Data}

\author{Zann~Koh,~Yuren~Zhou*,~\IEEEmembership{Member,~IEEE},~Billy~Pik~Lik~Lau,~\IEEEmembership{Student~Member,~IEEE}~Chau~Yuen,~\IEEEmembership{Senior~Member,~IEEE},~Bige~Tun\c cer,~and~Keng~Hua~Chong
\thanks{*Corresponding author}
\thanks{Zann~Koh, Yuren~Zhou, Billy~Pik~Lik~Lau, and  Chau~Yuen are with the Pillar of Engineering and Product Development, Singapore University of Technology and Design, Singapore. (e-mail: yuren\_zhou@sutd.edu.sg)}
\thanks{Bige~Tun\c cer and Keng~Hua~Chong are with the Pillar of Architecture and Sustainable Design, Singapore University of Technology and Design, Singapore.}}


\maketitle

\begin{abstract}
Information about the spatiotemporal flow of humans within an urban context has a wide plethora of applications. Currently, although there are many different approaches to collect such data, there lacks a standardized framework to analyze it. The focus of this paper is on the analysis of the data collected through passive Wi-Fi sensing, as such passively collected data can have a wide coverage at low cost. We propose a systematic approach by using unsupervised machine learning methods, namely $k$-means clustering and hierarchical agglomerative clustering (HAC) to analyze data collected through such a passive Wi-Fi sniffing method. We examine three aspects of clustering of the data, namely by time, by person, and by location, and we present the results obtained by applying our proposed approach on a real-world dataset collected over five months. 
\end{abstract}

\begin{IEEEkeywords}

passive Wi-Fi sensing, Wi-Fi sniffing, data mining, spatiotemporal, clustering.
\end{IEEEkeywords}

%
\IEEEpeerreviewmaketitle

\section{Introduction}

\IEEEPARstart{U}{nderstanding} the spatiotemporal flow of humans within urban areas is a useful undertaking as it has applications in fields such as urban planning\cite{zheng2011urban,becker2011tale}, crowd monitoring\cite{mowafi2013tracking}, and targeted advertising\cite{zhang2017targeted}. Such tracking of human flow can come under two categories: active and passive. Active tracking involves obtaining data from participants who are actively engaged in the data collection process. One difficulty present in the use of such active data collection methods is that many countries have policies put into place to protect the privacy of users, which makes such data unobtainable from apps without users' consent. On top of these policies, it is difficult to obtain such consent and active participation from a large number of people. 

In contrast to active tracking, passive tracking collects information about people without the need for active participation. Under the category of passive tracking, some currently used methods include cellular activity tracking \cite{gonzalez2008understanding,becker2013human,jiang2017activity,zhang2017understanding}, and the use of cameras\cite{ma2008areal,eshel2010tracking,kurilkin2015evaluation}. However, these methods may have certain drawbacks under certain conditions. For instance, the resolution of cellular methods depends on the density of cell towers in a region, which is usually in the scale of kilometers and would be too coarse to measure micro-mobility of humans. Camera-based tracking is usually limited to sparse crowds\cite{ma2008areal} and a small coverage area \cite{eshel2010tracking}. As limitations based on line-of-sight will be inherent in all camera-based tracking methods, we will look into using a different method of tracking human micro-mobility to avoid this. Alternative methods of tracking passively include the passive sniffing of Bluetooth and Wi-Fi signals from mobile phones.

Passive sniffing of Bluetooth signals has been used in contexts such as detecting human behavior in shopping mall environments\cite{galati2010human} as well as in public spaces such as the Louvre\cite{yoshimura2014analysis}. However, Bluetooth devices mainly operate in the range of 10 meters or so. This upper limit causes passive Bluetooth sensing to be less feasible for coverage of a large area such as a residential estate, as it would theoretically require the placement and maintenance of a large number of sensors throughout the course of the study. With Wi-Fi signals being detectable at a comparatively larger range of tens of meters, passive sensing of Wi-Fi signals was thus chosen as the method to focus on for this paper.

Within the available literature which makes use of passively detected Wi-Fi signals to make mobility-related inferences, there are some common methods of analysis. Most of them present the counts of detected devices over time, as well as heatmaps superimposed over maps of the region of detection. Some works \cite{nunes2017beanstalk,traunmueller2017digital,traunmueller2018digital} also have illustrations of the strength of each direct connection between sensors as a form of analysis of flow between each sensor.

However, previous literature on the use of Wi-Fi signals in mobility tracking lack the in-depth use of machine learning methods to examine passively collected Wi-Fi data in detail. The main challenges and research gap are as follows:
\begin{itemize}
	\item Lack of labels for Wi-Fi data - as passively collected Wi-Fi data is collected without the subjects' knowledge, there is no way to verify the accuracy of each person's time and location data with the person themselves.
	\item Lack of systematic approach in previous literature - in previous literature, there is a lack of a standardized systematic approach to address such passively collected Wi-Fi data.
	\item Noise present in Wi-Fi data - there will be noise present in real-world data when collecting it, therefore we have to find some way to clean it in order to extract useful insights.
\end{itemize}

In this paper, different from previous literature, we use clustering, a type of unsupervised machine learning technique, to analyze a large real-world dataset and thus obtain more insights than those obtained through the above methods alone. Clustering allows us to find groups of similar patterns within unlabeled data. As will be shown in the following sections, when applying clustering to Wi-Fi sensing data, one can obtain detailed multi-angle patterns related to people's behaviors in a certain residential or commercial district, which can be very helpful for district design and management. Examples include the clustering by time in Section IV, where the results can be used to check the similarities of people flow in different buildings across different days. This can help to inform the planning decisions of large-scale events in different buildings to avoid congestion, or to organize a roadshow during certain periods of time to capture a large number of passersby. The results of clustering by person as described in Section V can break down the population into a few groups with distinct behaviors, which can possibly have applications in marketing. Lastly, clustering by location as described in Section VI can help local authorities or urban planners detect whether the facilities within a residential estate are being utilized as planned, as well as track common routes of human travel within an estate for applications in future residential estate projects.

Therefore, this paper has the following contributions:
\begin{itemize}
  \item We propose a systematic approach which applies the unsupervised machine learning techniques, namely $k$-means clustering and hierarchical agglomerative clustering (HAC) to cluster a dataset gathered via the low-cost means of passive Wi-Fi sensing.
  \item We propose to analyze the data in three aspects - by time, by person, and by location.
  \item Finally, we apply the proposed approach on a real-world dataset collected over a period of five months and covering an area of approximately 0.52 km$^2$, and provide a detailed analysis of the clustering results.

\end{itemize}

The structure of the remaining sections will be as follows: Section II will discuss related works in detail. Section III will briefly go through the steps taken for data collection and preprocessing of the obtained data. Sections IV to VI will describe and present the results of clustering the data in three aspects - by time, by person, and by location respectively. Finally, Section VII will conclude the paper. 

\section{Related Works}
This section gives an overview of the related works divided into the subsections of mobility tracking methods and clustering methods.
\subsection{Tracking Methods}
Methods of active tracking of human mobility actively involve participants whose data is being tracked. Examples can be found in \cite{van2009sensing}, which used specific GPS sensing devices in tandem with interviews and questionnaires; in \cite{marakkalage2018understanding}, where the mobility and demographic data of participants were collected through the use of a mobile application; as well as in \cite{hu2016tales}, where geotagged data from a social media app was used.

However, the consent of participants is difficult to obtain, especially due to privacy concerns and legislation. This is evidenced by \cite{yip2016exploring}, in which participants were asked through letters to run an app for seven days which would log their GPS location, and they were required to input locations where they stayed for over half an hour. That experiment had a 1.33\% participation rate (45 individuals consented out of 3380 requests sent out). In addition, due to high reliance on individual users' compliance in self-reporting, data obtained through active methods may have some intrinsic bias and is likely to be scarcer than desired. Thus, for this study, we have chosen to use a passive tracking method.

Other than active tracking methods, there are also methods to passively track human mobility, such as cellular data tracking using cell towers\cite{gonzalez2008understanding,becker2013human,jiang2017activity,zhang2017understanding}, the use of cameras\cite{ma2008areal,eshel2010tracking,kurilkin2015evaluation}, and passive sensing of mobile phone signals such as Bluetooth and Wi-Fi.

While cell towers are readily present in the infrastructure of the country and the user base for collection of data is large, the data collected is difficult to access as it requires going through the service providers. Cell tower data also has a granularity in  the range of kilometers, which is too coarse to examine the micro-mobility of humans. As for camera-based tracking, it is mainly limited to sparse crowds \cite{ma2008areal} and a small coverage area\cite{eshel2010tracking}. Compounded with the inability to estimate the current distribution of people flowing into and out of each segment of the video\cite{kurilkin2015evaluation}, it is less feasible for use in our current study, which aims to study the micro-mobility of people within a residential estate and the nearby set of buildings, which would theoretically require a considerably large number of cameras to cover the entire area. 

We then turn to passive sensing of mobile phone signals. When making a decision between the use of Wi-Fi signals as compared to Bluetooth, we consider that it is more likely for a given device to have its Wi-Fi on as compared to its Bluetooth \cite{schauer2014estimating}. Sensing of Wi-Fi signals have been used in previous literature and works for finer resolution as compared to cell towers, as detection of packets is limited to the radius of Wi-Fi detection, which is typically less than 100 meters. Additionally, the authors in \cite{sapiezynski2015tracking} have shown that it is possible to infer a large proportion of mobility of a population from a time-series of Wi-Fi signals coupled with a small number of GPS data samples. This supports the use of passively collected Wi-Fi signals as a viable method of location detection over time.

In the previous literature, there have been a few works which authors have used sensors that were mobile, for example mobile phones held by volunteers \cite{chon2014sensing}, or laptops held by researchers themselves\cite{barbera2013signals}. The use of these sensors are manpower intensive and therefore become less feasible for data collection of daily mobility over a long period of time. For the remaining majority of works that have used sensing of Wi-Fi signals, the sensors used mostly consist of Wi-Fi access points or self-implemented infrastructures, which are stationary over the course of the data collection. This means that they do not require volunteer participation, and thus are ideal for passive people counting and tracking. 

Out of these papers that use stationary Wi-Fi sensors for their tracking, some mainly collect data for specific events rather than for daily human movement \cite{basalamah2016crowd,mcauley2017towards,alessandrini2017wifi,zhou2020understanding}; some are directed toward specific industries' applications such as investigating the movements within a hospital\cite{ruiz2014analysis}, within a university\cite{kalogianni2015passive}, within a shopping mall\cite{shen2018snow,shen2020bag}, within a public district with shopping malls and office buildings \cite{zhou2018understanding}, and among popular tourist locations on a set of islands \cite{nunes2017beanstalk}; and a few cover large enough parts of cities such as Manhattan and Copenhagen to identify the daily movement of commuters between work and home\cite{traunmueller2017digital,traunmueller2018digital,sapiezynski2015tracking}. Although the scopes of these papers vary widely, they have certain similarities in the analysis of the data.  Most of them present the counts of detected devices over time, as well as heatmaps superimposed over maps of the region of detection. Some works \cite{nunes2017beanstalk,traunmueller2017digital,traunmueller2018digital} also have illustrations of the strength of each direct connection between sensors as a form of analysis of flow between each sensor. However, these previous literature all lack the systematic use of machine learning methods to gather deeper insights into their mobility data. Therefore, this paper will propose the use of a machine learning method, clustering, to discover common mobility patterns from the data we have collected.

\subsection{Clustering Methods}
Within the field of machine learning, there are two main types: supervised learning, which requires labeled data, and unsupervised learning, which does not. In this case, as the data we have collected lacks ground truth, we will be using an unsupervised machine learning method, more specifically clustering. Clustering methods find groups in the input data based on input parameters and a distance measure.

As our data size is large, we would require the algorithms we use to be scalable. Some algorithms are not scalable, such as Mean Shift clustering \cite{comaniciu2002mean}, which requires multiple nearest neighbor searches during the execution of the algorithm, and Balanced Iterative Reducing and Clustering using Hierarchies (BIRCH) \cite{zhang1996birch}, which does not scale well to high dimensional data. Thus, these algorithms are less ideal for this study.

Other clustering algorithms such as Density-Based Spatial Clustering of Applications with Noise (DBSCAN) \cite{ester1996density} and Ordering Points to Identify the Clustering Structure (OPTICS) \cite{ankerst1999optics} are computationally more complex and requires more complicated, potentially iterative parameter selection. Additionally, as we hope to potentially discover unusual mobility patterns, it would be more difficult to tune DBSCAN's parameters to obtain explainable and meaningful clusters, as compared to k-means.
	
The $k$-means clustering algorithm\cite{lloyd1982least}, on the other hand, has the advantages of being simple to implement, with a single intuitive choice of parameter (number of clusters, $k$), as well as being scalable to large datasets. The $k$-means clustering algorithm has been extensively studied in cases with outliers as well, and it has been shown in works such as those by Im et al \cite{im2020fast} and Bhaskara et al \cite{bhaskara2019greedy} that even in datasets with noise, a high-quality clustering can be obtained after removal of outliers, while using the $k$-means algorithm. Other literature on using k-means on a noisy dataset can also be found in the references \cite{tang2004noise,ben2014clustering,manjoro2016operational,schelling2018kmn,he2019clustering}. In addition to $k$-means, hierarchical agglomerative clustering (HAC)\cite{mullner2011modern} is also a possible choice of algorithm, as it is also scalable to large datasets and has an intuitive visualization in terms of the dendrogram. Users are able to easily view the distance between each level of agglomerated clusters and decide on an appropriate distance threshold. Thus, for this paper, we will be using the $k$-means clustering algorithm as well as the HAC algorithm for the purposes of clustering.

\section{Data Collection and Preprocessing}
\subsection{Data Collection}

Passive Wi-Fi sensors were deployed at selected locations of interest in two areas of the city: a Facility area and a Residential area. The Facility area consists of a hospital building, an educational institution, and four shopping mall buildings containing offices. For ease of explanation, the shopping mall buildings will be referred to as Malls 1 to 4, respectively. Malls 3 and 4 are located near a transportation hub, which is hypothesised to be the main means of transport to and from the Facility area from other places outside. All of the buildings in the Facility area are connected with an elevated walkway, and our sensors were installed at the entrances to the buildings from the walkways. The Residential area has 38 blocks in total, and our sensors were deployed at 14 selected locations among these buildings. The distance between these 2 areas is in the range of a few hundred metres, and this Facility area would be the nearest commercial center to the Residential area. Sketches of the different buildings in each area can be found in Fig. 8 for a clear understanding. 

\begin{figure}
    \centering
    \includegraphics[width=\linewidth]{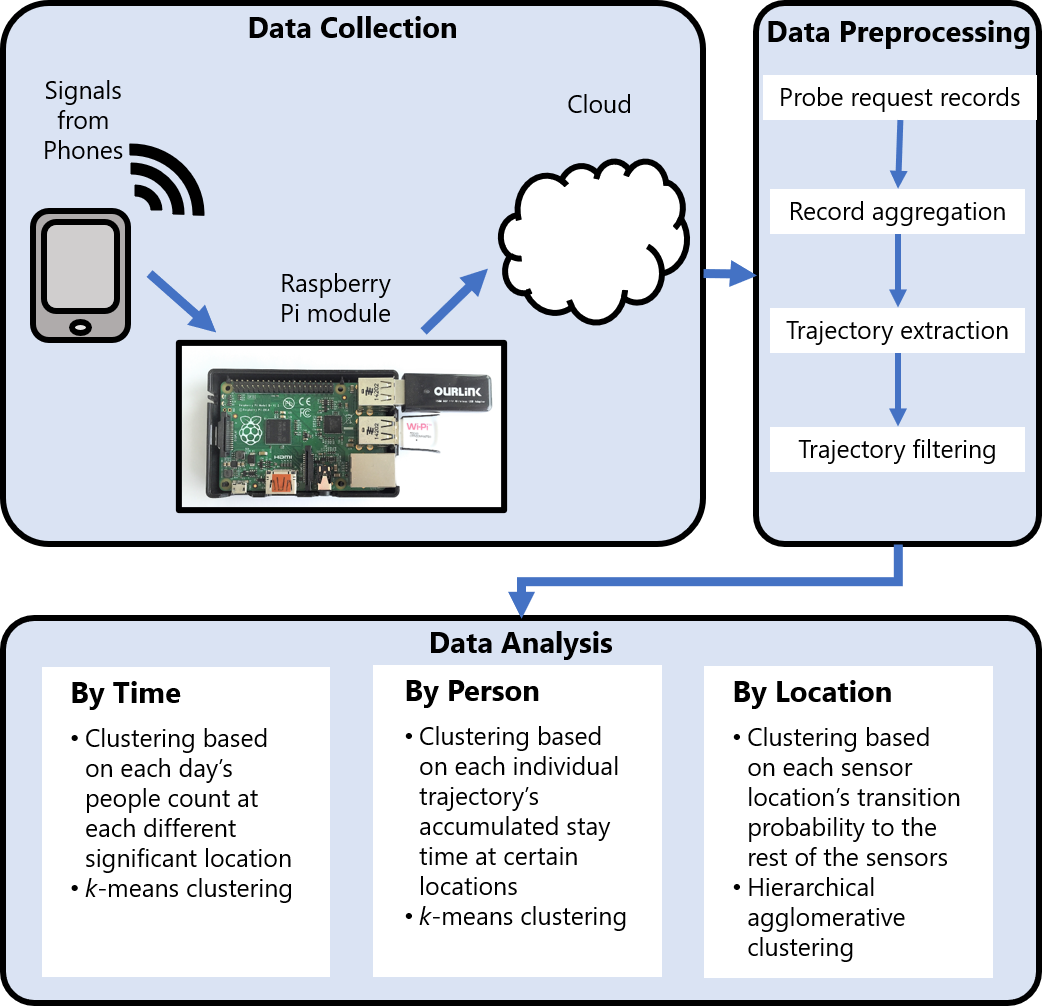}
    \caption{\textit{Overall framework for understanding spatiotemporal human flow through passive Wi-Fi sensing and mining of collected data.}}
    \label{fig:flowchart}
\end{figure}

The sensors passively collect Wi-Fi probe packets sent from nearby mobile devices. Each sensor is a box containing a Wi-Fi sniffer, which is built on top of the Raspberry Pi Model B with additional WI-PI USB dongle for probe request collection. Local processing of probe requests is performed to reduce the volume of uploaded data, in turn reducing costs of data transmission and storage. This local processing involves combining probe requests with a chronological separation of 3 minutes or less \cite{li2018experimental}. The processed data is then uploaded to the cloud via cellular connection. A flowchart illustrating the above process can be found in Fig.~\ref{fig:flowchart}.

\subsection{Preprocessing}

The collected data has to be further processed using
the method in our earlier work \cite{li2018experimental} to obtain the trajectory of a particular mobile device with a unique MAC address. The trajectories obtained using the above method are sensor level trajectories. For this study, we will do some merging of the sensor level trajectories to obtain building level trajectories, which will be explained later. After the trajectories are obtained, due to the noisy nature of such data collected using Wi-Fi probes, further filtering based on heuristics is performed as described later on in this section.

A sensor level trajectory is represented by ($x_1,x_2,...,x_n$), in which $x_i$ = ($macAddress$, $sensorID_i$, $nextSensor_i$, $startTime_i$, $endTime_i$, $stayTime_i$, $takeTime_i$). As each device's trajectory is grouped together by MAC address, the value for $macAddress$ for each $x_i$ in the same trajectory will be the same. No other information about devices was retained, which keeps the privacy of device owners from being compromised as we cannot use the MAC address alone to identify a specific individual. $sensorID_i$ and $nextSensor_i$ belong to the set of all sensors used in this study, $\{A1, A2, B1, B2, ...\}$. The first character of each sensor name refers to each building where the sensor was placed, while the remaining digits serve to differentiate different sensors placed at the same building. For example, A1 and A2 represent two different sensors, both placed at the building labeled `A'. In the code, the first character of $sensorID_i$ is a string variable, so the building is represented by the first character of the string, $sensorID_i$[0]. For example, if $sensorID_i$ is `A1', $sensorID_i$[0] would be `A'. $sensorID_i$ represents the ID of the $i$th sensor in the trajectory, while $nextSensor_i$ represents the ID of the $(i + 1)$th sensor. 

For the last element in the trajectory $x_n$, $nextSensor_n$ will be left empty. $startTime_i$ and $endTime_i$ represent the start time and end time of detection at sensor $i$ respectively. $stayTime_i$ (time spent at the current sensor) and $takeTime_i$ (time taken to reach the next sensor after leaving current sensor) are calculated as in Eqn.~(\ref{eq:2}) and (\ref{eq:3}) respectively:

\begin{equation}
stayTime_i = endTime_i - startTime_i
\label{eq:2}
\end{equation}

\begin{equation}
\scalebox{.85}{$
takeTime_i =
\begin{cases}
startTime_{i+1} - endTime_i & \text{if $i = 1,...,n-1$} \\
0 & \text{if $i = n$}
\end{cases}$}
\label{eq:3}
\end{equation}

Each trajectory lasts for the length of one day. Days were taken from 3:00 AM on one calendar day to 3:00 AM the next, to give allowance for some commercial or social activities that may cut across midnight. 

\begin{algorithm}[ht]
	\caption{Merging of sensor level trajectory}
	\begin{algorithmic}
		\Require $Traj_{input}$: original sensor level trajectory
		\Ensure $Traj_{new}$: merged building level trajectory
		\Statex
		\State $n \gets length(Traj_{input})$
		\State $Traj_{new} \gets []$
		\For{$i = 1,...,n$} \Comment{Change sensor names to buildings}
		\State $sensorID_i \gets sensorID_i[0]$
		\State $nextSensor_i \gets nextSensor_i[0]$
		\EndFor
		\For{$i = 1,...,n$}
		\If{$takeTime_i<21600$ and $nextSensor_i=sensorID_i}$
		\State $macAddress_{new} \gets macAddress_i$
		\State $sensorID_{new} \gets sensorID_i$
		\State $nextSensor_{new} \gets nextSensor_{i+1}$
		\State $startTime_{new} \gets startTime_i$
		\State $endTime_{new} \gets endTime_{i+1}$
		\State $stayTime_{new} \gets$ 
		\State $(stayTime_i+takeTime_i+stayTime_{i+1})$
		\State $takeTime_{new} \gets takeTime_{i+1}$
		\State $x_{new} \gets (macAddress_{new}, sensorID_{new},$ 
		\State $nextSensor_{new},startTime_{new}, endTime_{new},$ 
		\State $stayTime_{new},takeTime_{new})$
		\Else
		\State $x_{new} \gets x_i$
		\EndIf
		\State $Traj_{new} \gets [Traj_{new},x_{new}]$
		\EndFor
	\end{algorithmic}
	
\end{algorithm}

For the purpose of this study, we consider trajectories at building level instead of sensor level, as each individual building has its own purpose (such as shopping mall, hospital etc) and thus any results we obtain could be explained more intuitively as compared to individual sensor levels. To get the building level trajectories from sensor level trajectories, consecutive entries detected at the same building in the same trajectory were merged if they occurred within 6 hours of each other on the same day according to Algorithm 1. Two sensors are from the same building if the first character in the sensor name is the same, such as A1 and A2.  

An example of merging a trajectory is shown in Fig. 2. It can be seen that the detections at Buildings A and C were merged, but the detections at Building B were not, due to exceeding the time threshold.

Other than merging, all trajectories with $endTime_n - startTime_1$ ($n$ being the trajectory length) less than 5 minutes were discarded as these short trajectories are not informative and are therefore outside the scope of our investigation. Long trajectories indicating that the device stayed at a single location for more than 16 hours in a single day were similarly discarded as they are deemed to be anomalies. After this filtering, our dataset consisted of more than 5.7 million trajectories from around 1.6 million devices over the period of 5 months.

\begin{figure}[t]
	\centering
	\includegraphics[width=\linewidth]{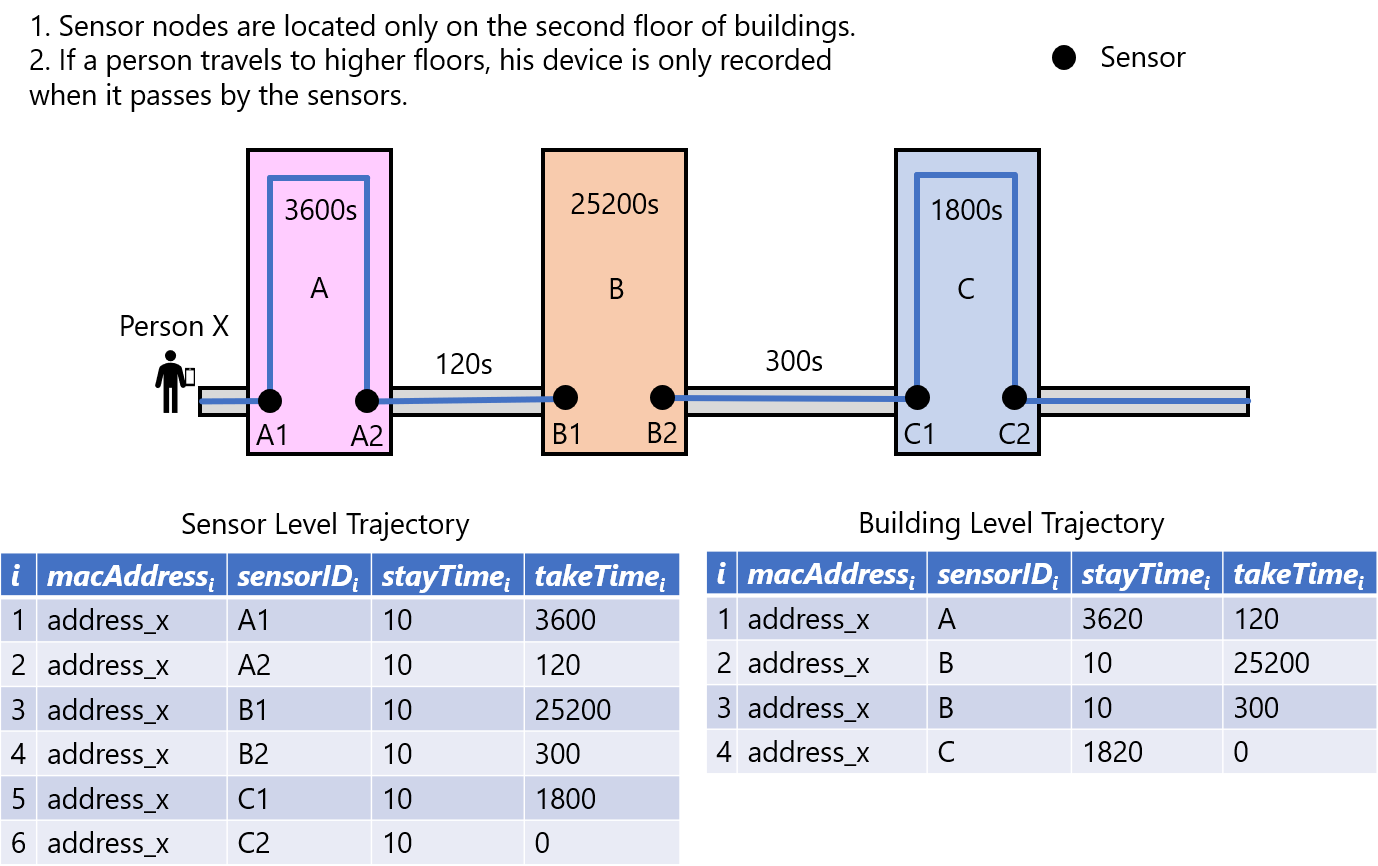}
	\caption{\textit{An example of merging certain consecutive entries along a trajectory.}}
	\label{fig:bldgs}
\end{figure}

\section{Clustering By Time}

This section explores the first aspect of trajectory data clustering, which is clustering by time. This section will be split into two parts: the results of the clustering of calendar days according to each day's features at each location, as well as some further analysis of highlighted clustering results.

\subsection{Clustering Results}
Clustering by time involves the extraction of features into a vector representing each day and clustering those vectors. Each day is segmented into 24 intervals of 1 hour each, starting from 3:00 AM on one day to 3:00 AM the next day. The number of people appearing at each location within that hour was extracted from the data, and this forms a 1-by-24 feature vector for each day. Each vector was normalized using min-max normalization before being subjected to $k$-means clustering. The aim of this clustering is to investigate the possibility of inferring temporal context such as type of day in terms of weekday as compared to weekend based on the patterns of people count at each location in the Facility area. The $k$-means algorithm is chosen based on its scalability to large datasets and simplicity in choice of a single parameter. 

The $k$-means algorithm uses expectation-minimization to perform clustering as described in Eqn. \ref{eq:4} to \ref{eq:5} below. Firstly, $k$ centroids $C_j$ where $j$ = 1,...,$k$ are initialized within the feature space of the data set. Second, each data point is temporarily assigned with an index equal to the index of the centroid nearest to it. The index of the $i$th data point $x_i$ out of the total number of $N$ data points is referred to as $ind_i$ ($ind_i \in \{1,...,k\}$) and is computed using Eqn. \ref{eq:4}. The distance metric commonly used is Euclidean distance. Next, the centroids of each cluster are reassigned to the arithmetic mean of all the data points in each cluster as in Eqn. \ref{eq:5}, where $n_j$  refers to the number of data points in cluster $j$. The last two steps are then repeated until convergence, which occurs when the calculated means of the clusters do not change in subsequent steps. 

\begin{equation}
	ind_i := arg \min_j ||x_i - C_j||^2
	\label{eq:4}
\end{equation}

\begin{equation}
	C_j := \frac{1}{n_j}\sum_{i:ind_i = j} x_i
	\label{eq:5}
\end{equation}

\begin{figure}[ht]
	\centering
	\includegraphics[width=0.75\linewidth]{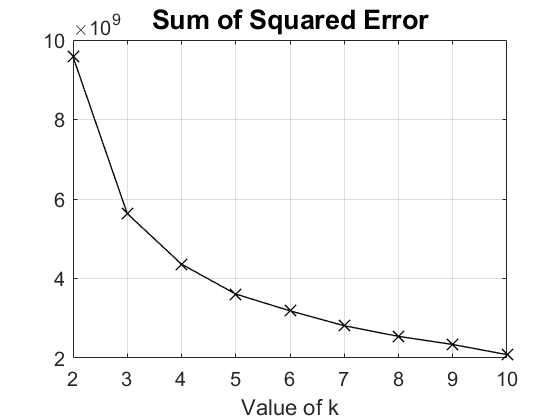}
	\caption{\textit{Plot of the sum-of-squared error values for different values of $k$.}}
	\label{fig:sse}
\end{figure}

This is a fast and simple method to cluster data, and it works well for data with features of a similar type - in this case, all features are numbers of people detected at a location within a certain hour. To find the best value of $k$ to set as a parameter for this clustering, we plotted the sum-of-squared error plot as shown in Fig.~\ref{fig:sse}, which indicates that the ideal value for $k$, located at the elbow point of the curve, is between 3 and 5. After testing out these values for $k$, a value of $k$ = 4 turned out to give the best balance between specificity and interpretability, as a value of 3 could be broken down further whereas the value of 5 broke down the clusters in an unintuitive way. From the results, it also largely corresponds to 4 main types of days with their own characteristics - working weekdays, Fridays, Saturdays and Sundays.

\begin{figure}[ht]
    \centering
    \includegraphics[width=\linewidth]{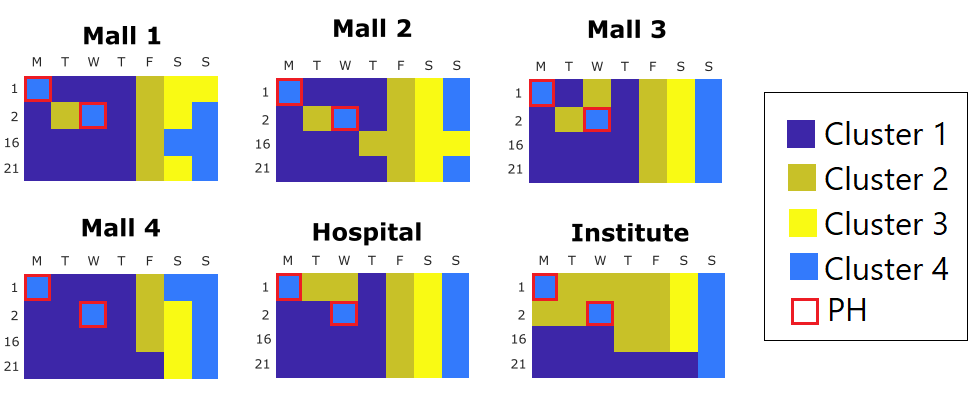}
    \caption{\textit{Representation of results of clustering by time. Weeks 1, 2, 16, and 21 are selected for clarity. Weeks 1 and 2 show public holidays (PH) that are grouped into the same cluster as Sundays, while weeks 16 and 21 are selected as an example of 'normal weeks' without PH or special days.}}
    \label{fig:calplots_all}
\end{figure}

Fig.~\ref{fig:calplots_all} shows a plot of the results of clustering by day in calendar form. In each calendar, the column represents one day of the week from Monday through Sunday, while each row represents the week number of the data, with week 1 being the first week of data collection, and so on.

Weeks 1 and 2 were selected to show the cluster assignment of public holidays (PH) and PH eves. The PH in weeks 1 and 2 are the squares with red borders. Weeks 16 and 21 were chosen to show the cluster assignment of a "normal" week, that is, without special days like PH or PH eves. 

From the calendars of Malls 1 to 4 and the Hospital, it can be seen that the clustering generally follows the four types of days below with overall at least 70\% of instances falling into each respective cluster: Cluster 1 (dark blue) mainly has working Mondays to Thursdays, Cluster 2 (brown) mainly contains Fridays, Cluster 3 (yellow) mainly contains Saturdays, and Cluster 4 (light blue) mainly contains Sundays. As seen from Table 1(a) to 1(f), 100\% of the available PH in the dataset are clustered in the same cluster as Sundays (Cluster 4), while PH eves are clustered in the same cluster as Fridays (Cluster 2) in a proportion of 50\% and above for five out of the six buildings. 
\begin{table*}
	\def\arraystretch{1.5}
	\fontsize{8pt}{8pt}\selectfont\centering
	\caption{Percentage of actual (row) vs clustered (column) days in each separate building}
	\begin{tabular}{cc}
		\multirow{2}{*}{\textbf{(a) Hospital}} & \multirow{2}{*}{\textbf{(b) Institute}}\\
		\\
		\def\arraystretch{1.5}
		\begin{tabular}{|c|c|c|c|c|c|c|}
			\hline
			& \textbf{Mon-Thur} & \multicolumn{2}{c|}{\textbf{Fri/PH Eve}} & \textbf{Sat} & \multicolumn{2}{c|}{\textbf{Sun/PH}}\\
			\hline
			\textbf{Mon-Thur} & 51.4 & 48.6 & 0.0 & 0.0 & 0.0 & 0.0\\
			\hline
			\textbf{Fri} & 16.7 & 83.3 & 0.0 & 0.0 & 0.0 & 0.0\\
			\hline
			\textbf{PH Eve} & 100.0 & 0.0 & 0.0 & 0.0 & 0.0 & 0.0\\
			\hline
			\textbf{Sat} & 0.0 & 0.0 & 0.0 & 100.0 & 0.0 & 0.0\\
			\hline
			\textbf{Sun} & 0.0 & 0.0 & 0.0 & 0.0 & 100.0 & 0.0\\
			\hline
			\textbf{PH} & 0.0 & 0.0 & 0.0 & 0.0 & 0.0 & 100.0\\
			\hline
		\end{tabular} &
		\def\arraystretch{1.5}
		\begin{tabular}{|c|c|c|c|c|c|c|}
			\hline
			& \textbf{Mon-Thur} & \multicolumn{2}{c|}{\textbf{Fri/PH Eve}} & \textbf{Sat} & \multicolumn{2}{c|}{\textbf{Sun/PH}}\\
			\hline
			\textbf{Mon-Thur} & 51.4 & 42.9 & 0.0 & 2.9 & 2.9 & 0.0\\
			\hline
			\textbf{Fri} & 35.0 & 65.0 & 0.0 & 0.0 & 0.0 & 0.0\\
			\hline
			\textbf{PH Eve} & 0.0 & 0.0 & 100.0 & 0.0 & 0.0 & 0.0\\
			\hline
			\textbf{Sat} & 11.1 & 0.0 & 0.0 & 72.2 & 16.7 & 0.0\\
			\hline
			\textbf{Sun} & 0.0 & 0.0 & 0.0 & 21.4 & 78.6 & 0.0\\
			\hline
			\textbf{PH} & 0.0 & 0.0 & 0.0 & 0.0 & 0.0 & 100.0\\
			\hline
		\end{tabular} \\
		\multirow{2}{*}{\textbf{(c) Mall 1}} & \multirow{2}{*}{\textbf{(d) Mall 2}}\\
		\\
		\def\arraystretch{1.5}
		\begin{tabular}{|c|c|c|c|c|c|c|}
			\hline
			& \textbf{Mon-Thur} & \multicolumn{2}{c|}{\textbf{Fri/PH Eve}} & \textbf{Sat} & \multicolumn{2}{c|}{\textbf{Sun/PH}}\\
			\hline
			\textbf{Mon-Thur} & 71.8 & 26.9 & 0.0 & 0.0 & 1.3 & 0.0\\
			\hline
			\textbf{Fri} & 4.8 & 95.2 & 0.0 & 0.0 & 0.0 & 0.0\\
			\hline
			\textbf{PH Eve} & 0.0 & 0.0 & 100.0 & 0.0 & 0.0 & 0.0\\
			\hline
			\textbf{Sat} & 0.0 & 0.0 & 0.0 & 76.2 & 23.8 & 0.0\\
			\hline
			\textbf{Sun} & 0.0 & 0.0 & 0.0 & 11.1 & 88.9 & 0.0\\
			\hline
			\textbf{PH} & 0.0 & 0.0 & 0.0 & 0.0 & 0.0 & 100.0\\
			\hline
		\end{tabular} &
		\def\arraystretch{1.5}
		\begin{tabular}{|c|c|c|c|c|c|c|}
			\hline
			& \textbf{Mon-Thur} & \multicolumn{2}{c|}{\textbf{Fri/PH Eve}} & \textbf{Sat} & \multicolumn{2}{c|}{\textbf{Sun/PH}}\\
			\hline
			\textbf{Mon-Thur} & 94.7 & 4.0 & 0.0 & 1.3 & 0.0 & 0.0\\
			\hline
			\textbf{Fri} & 0.0 & 100.0 & 0.0 & 0.0 & 0.0 & 0.0\\
			\hline
			\textbf{PH Eve} & 50.0 & 0.0 & 50.0 & 0.0 & 0.0 & 0.0\\
			\hline
			\textbf{Sat} & 0.0 & 0.0 & 0.0 & 100.0 & 0.0 & 0.0\\
			\hline
			\textbf{Sun} & 0.0 & 0.0 & 0.0 & 16.7 & 83.3 & 0.0\\
			\hline
			\textbf{PH} & 0.0 & 0.0 & 0.0 & 0.0 & 0.0 & 100.0\\
			\hline
		\end{tabular} \\
		
		\multirow{2}{*}{\textbf{(e) Mall 3}} & \multirow{2}{*}{\textbf{(f) Mall 4}}\\
		\\
		\def\arraystretch{1.5}
		\begin{tabular}{|c|c|c|c|c|c|c|}
			\hline
			& \textbf{Mon-Thur} & \multicolumn{2}{c|}{\textbf{Fri/PH Eve}} & \textbf{Sat} & \multicolumn{2}{c|}{\textbf{Sun/PH}}\\
			\hline
			\textbf{Mon-Thur} & 79.5 & 20.5 & 0.0 & 0.0 & 0.0 & 0.0\\
			\hline
			\textbf{Fri} & 9.5 & 90.5 & 0.0 & 0.0 & 0.0 & 0.0\\
			\hline
			\textbf{PH Eve} & 50.0 & 0.0 & 50.0 & 0.0 & 0.0 & 0.0\\
			\hline
			\textbf{Sat} & 0.0 & 0.0 & 0.0 & 95.2 & 4.8 & 0.0\\
			\hline
			\textbf{Sun} & 0.0 & 0.0 & 0.0 & 0.0 & 100.0 & 0.0\\
			\hline
			\textbf{PH} & 0.0 & 0.0 & 0.0 & 0.0 & 0.0 & 100.0\\
			\hline
		\end{tabular} &
		\def\arraystretch{1.5}
		\begin{tabular}{|c|c|c|c|c|c|c|}
			\hline
			& \textbf{Mon-Thur} & \multicolumn{2}{c|}{\textbf{Fri/PH Eve}} & \textbf{Sat} & \multicolumn{2}{c|}{\textbf{Sun/PH}}\\
			\hline
			\textbf{Mon-Thur} & 83.1 & 16.9 & 0.0 & 0.0 & 0.0 & 0.0\\
			\hline
			\textbf{Fri} & 5.3 & 94.7 & 0.0 & 0.0 & 0.0 & 0.0\\
			\hline
			\textbf{PH Eve} & 50.0 & 0.0 & 50.0 & 0.0 & 0.0 & 0.0\\
			\hline
			\textbf{Sat} & 0.0 & 0.0 & 0.0 & 90.0 & 10.0 & 0.0\\
			\hline
			\textbf{Sun} & 0.0 & 0.0 & 0.0 & 17.6 & 82.4 & 0.0\\
			\hline
			\textbf{PH} & 0.0 & 0.0 & 0.0 & 0.0 & 0.0 & 100.0\\
			\hline
		\end{tabular}

	\end{tabular}	
\end{table*}

Now that we know that different types of days follow generally different patterns, the actual patterns of each day are then investigated in the next subsection.

\subsection{Further Analysis of Clustering Results}

\begin{figure}[ht]
	\centering
	\includegraphics[width=\linewidth]{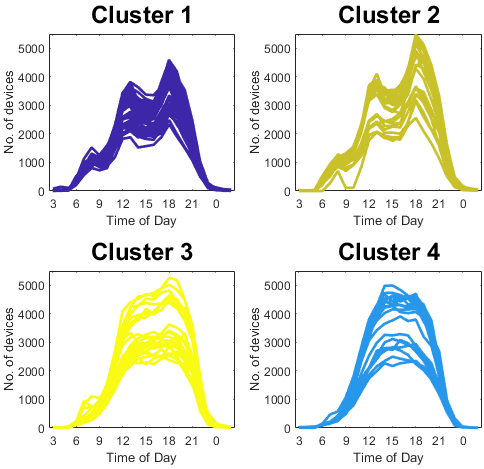}
	\caption{\textit{Curves of device count versus time at Mall 2 on different days, separated based on the clustering results. The four clusters are mainly corresponding to 1) Working Mondays to Thursdays, 2) Fridays/Public holiday eves, 3) Saturdays, and 4) Sundays/PH. The x-axis of each plot represents the time of day while y-axis represents normalized device count.}}
	\label{fig:backend}
\end{figure}

To examine how the actual patterns of people count vary in different clusters, an example of people count curves on different clusters of days from Mall 2 was visualised in Fig.~\ref{fig:backend}. The colors correspond to the cluster assignments in Fig.~\ref{fig:calplots_all}. 

Upon first glance, the curves for Clusters 1 (Mondays to Thursdays) and 2 (Fridays and PH Eves) are very similar to each other and very different from Clusters 3 (Saturdays) and 4 (Sundays and PH), which are also similar to each other. For Clusters 1 and 2, they each have three peaks of people count, which tend to occur between 6:00 AM to 9:00 AM, between 11:00 AM and 2:00 PM, and between 5:00 PM to 8:00 PM. These periods of time roughly correspond to office hours and meal times, so these could show the surges of people arriving at the mall to have meals within those time periods. The last peak that occurs in the evening could also represent the evening shoppers. However, the first peak of people count is lower for Cluster 2 than Cluster 1, and the last peak is much higher. There could be an increase in people going to the mall in the evenings on Friday/PH eve to shop as compared to working Mondays to Thursdays. 

For Clusters 3 and 4, the curves are generally smooth, increasing sharply in the late morning and plateauing across the middle of the day before dropping back down to zero after 10:00 PM. Cluster 3's curve gently increases from 1:00 PM to 7:00 PM, reaching its highest point at 7:00 PM, while Cluster 4's curve lacks a noticeable increase in the evening. This could be because people are more likely to stay out later in the evenings on Saturdays as they do not need to go to work the following morning, compared to Sundays when they do.

\section{Clustering By Person}
This section explores the second aspect of trajectory data clustering, which is clustering by person. This section will be split into two parts: the results of the clustering of individual trajectories, and the temporal analysis of highlighted clustering results based on the analysis of the above section. 

\subsection{Clustering Results}
Individual trajectories within a single day had their features extracted into a single vector. The features used for clustering were the accumulated time detected in a single day at the hospital, shopping mall, educational institute, the residential estate (day time), and the residential estate (night time) respectively. The detected time at the residential estate was split into between 7:00 AM to 7:00 PM (day time) and 7:00 PM to 7:00 AM (night time). This split was performed as many retired elderly were observed to be staying at the estate and they are hypothesized to be more active in the day time, as opposed to the working adults or school children, who are likely to be more active in the residential estate later in the day or evening. These features were then used as input for clustering via the $k$-means clustering algorithm. 

\begin{figure*}[!ht]
	\centering
	\includegraphics[width=\linewidth]{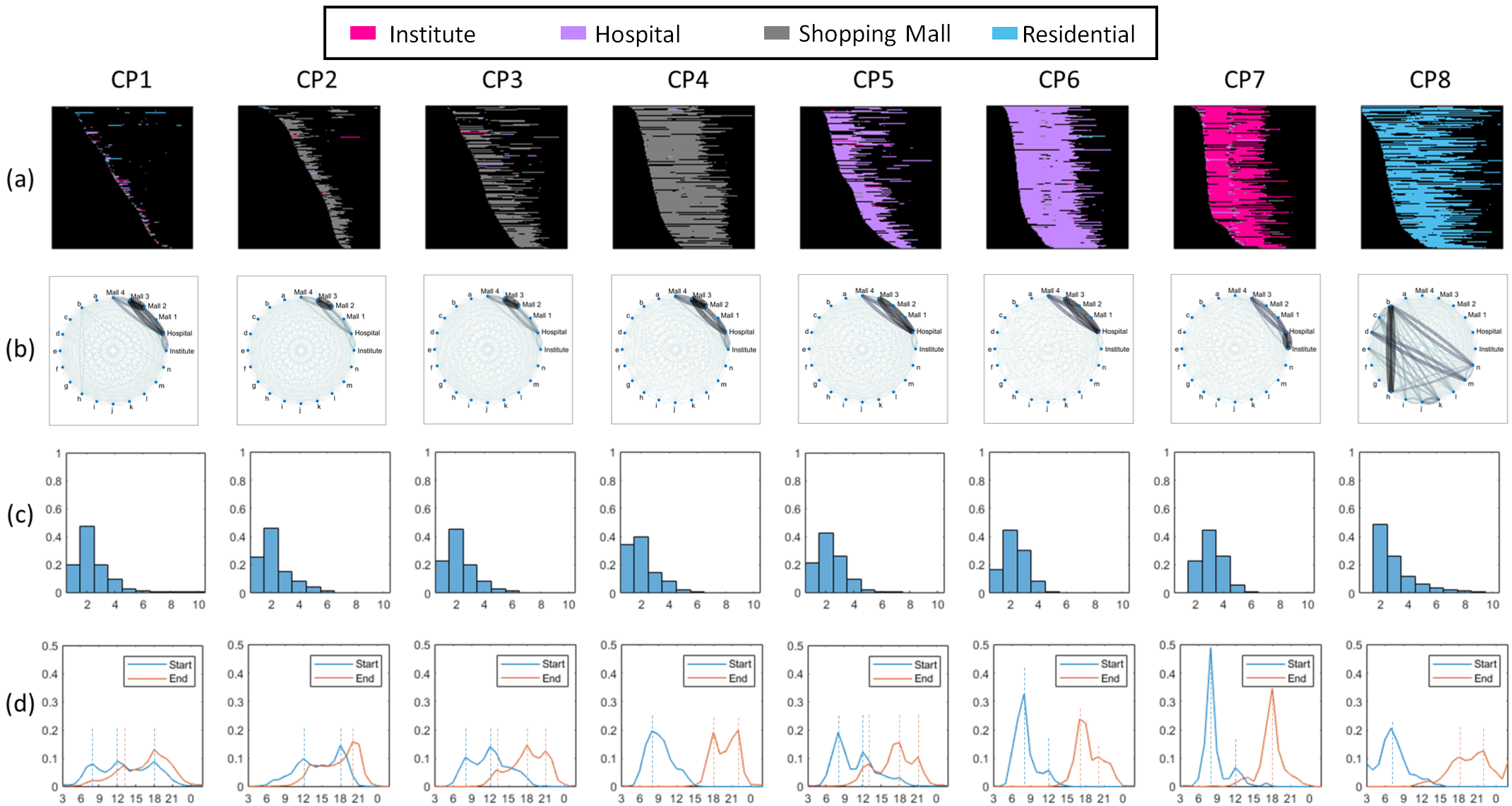}
	\caption{\textit{Illustrations of results from clustering data by person. (a) A visual representation of the eight clusters obtained using k-means clustering of trajectory stay times at differing locations. The horizontal axis represents the time of day from 3:00 AM to 3:00 AM the next day, while the color represents the location where the device was detected, as described in the legend. Trajectories are sorted by the time at which they were first detected. (b) Visualizations of transition probability between each pair of nodes within each cluster, for comparing the movement patterns between clusters. A darker and thicker line connecting a pair of nodes means that there are more transitions between those nodes. (c) Histograms describing the number of unique locations visited per trajectory within each cluster. The y-axis has been subjected to probability normalization. (d) Probability distributions of start times and end times of trajectories in each cluster. The x-axis represents the time of day and the y-axis represents the probability that a trajectory in the respective cluster starts or ends within that timing with a resolution of one hour.}}
	\label{fig:byperson}
\end{figure*}

A value for $k$ has to be predetermined for use in $k$-means. Traditional indices for calculating the optimal number of clusters for k-means clustering, such as the silhouette index\cite{rousseeuw1987silhouettes} and sum-of-squared error plot, suggests an ideal value of 5. However, when trying to use this value, the sizes of the clusters are greatly skewed. We therefore manually increase $k$ to find a suitable value for clustering and end up with a value of 8. When the value of $k$ was originally selected as 5, the people who stayed for a longer time at the shopping malls were grouped together to form a very large cluster. After the value of $k$ was increased to 8, this large cluster was further divided into 3 smaller clusters, each representing one more specific type of person. Similarly, the people who stayed the longest time at the hospital were originally grouped together in a large cluster in the $k$ = 5 case and this large cluster was divided into 2 smaller clusters after $k$ was increased to 8. As a result, $k$ = 8 is selected over 5 because it produces a more balanced and informative clustering result. This value was used to perform $k$-means clustering on the feature vectors. The aim of this portion is to search for clusters of device trajectories and, from there, infer insights about a given device based on the cluster that its trajectory is assigned to. The results of the clustering and each part of our proposed analysis is shown in Fig.~\ref{fig:byperson}. Each cluster is labeled as `CP', which stands for Cluster of People, together with its corresponding number.

Row (a) in Fig.~\ref{fig:byperson} contains the visualizations of the trajectories of the individual clusters. The colors represent the different types of buildings as stated in the legend.  Row (b) contains the plots of transition probabilities between each pair of nodes within each cluster, in order to study the movement patterns. There are 20 nodes in total, with 6 from the Facility area and 14 individual buildings from the Residential area, labeled alphabetically from 'a' to 'n'. Row (c) contains the histograms showing the distribution of number of unique nodes visited per trajectory in each cluster. Finally, Row (d) contains the probability distributions of start and end times of trajectories in each cluster. 

From row (a), these eight clusters can be broken down visually into 3 groups of similar clusters - CP1 to CP4 being mainly at the shopping malls (gray), CP5 and CP6 being mainly at the hospital (purple), CP7 being mainly at the Institute (pink), and CP8 being mainly at the Residential area (blue). Each cluster will be discussed further in detail.

CP1 to CP4 are the clusters in which most, if not all, trajectories have visited a shopping mall at least once. The amount of time stayed at a shopping mall per trajectory increases in order from CP1 (less than 1 hour) to CP4 (7.5 to 15 hours). The transition probability diagrams show heavy emphasis on the buildings in the Facility area (top right). The most common transitions are those between Malls 2 and 3, followed by the transitions between Mall 3 and the Hospital, as well as between Malls 2 and 4. All four of these clusters also show a similar pattern in terms of number of nodes visited per trajectory peaking at two, however in CP4 there is a much higher probability of a trajectory containing a single building as compared to the other three clusters. The main difference in these clusters lie in the distributions of start and end times. CP1 has three peaks for the start time (8:00 AM, 12:00 PM, 6:00 PM), while CP2 and CP3 have two peaks (12:00 PM and 6:00 PM for CP2, 8:00 AM and 12:00 PM for CP3), and CP4 has only one (8:00 AM). These common timings correspond with the common times to start work (8:00 AM), break for lunch (12:00 PM), and leave work for home (6:00 PM). Those people who appear in the shopping mall at 6:00 PM and stay there for at least an hour (CP2) could be going for short shopping trips after work. Those who appear at 8:00 AM may either pass by the shopping malls on their way to work (CP1) or start working at the shopping mall in the morning (CP3 and CP4). However, since the time spent ranges from between 3.5 and 7.5 hours (CP3) as compared to between 7.5 and 15 hours (CP4), it can be inferred that trajectories in CP4 are more likely to belong to people working in the shopping malls for long hours, while those in CP3 could belong to either people with shorter shifts, or long shopping trips.

CP5 and CP6 are the clusters in which all the trajectories have visited the Hospital and stayed for a relatively long time (2 to 6.5 hours for CP5 and 7.5 to 15 hours for CP6). The transition probability diagrams both show strong links between Mall 3 and Hospital, as well as Mall 4 and Hospital. Since Malls 3 and 4 are connected to a transport hub, these high number of transitions could reflect the people taking public transport to and from the Facility area. The distribution of the number of nodes visited per trajectory is also similar, both peaking at two places and decreasing with increasing number of places up till five or six. The main difference between these two clusters lies in the distributions of the start and end times. Both clusters have peaks in start time at 8:00 AM and 12:00 PM, however the 8:00 AM peak for CP6 is much higher and the 12:00 PM peak is much lower as compared to CP5. For the ending times, both clusters have peaks at 5:00-6:00 PM and 9:00 PM, but the 6:00 PM peak is much higher for CP6 than CP5. This indicates that the people in CP6 start and end their trajectories over a much narrower time period than CP5, which is in line with the consideration that people in CP6 may be workers at the Hospital, while people in CP5 may be either shift workers or visitors to the Hospital.

CP7 is the cluster that contains trajectories with the longest time at the Institute, from 3.5 to 14.5 hours. Its corresponding transition probability diagram is noticeably different from the others, with strong links between the Hospital and the Institute, as well as between Mall 3 and the Hospital. These strong links may also be explained as above, due to the proximity of Mall 3 to the transport hub. Most of the trajectories in this cluster contained three unique nodes, which is different from the rest as well, since the rest of the unique node distributions all peaked at two. The start and end time distributions are also unique, in that there is a single prominent sharp spike in each of the start and end times. The start time peak occurs at 8:00 AM with a probability value of almost 0.5, while the end time peak occurs at 6:00 PM with a value of between 0.3 to 0.4. This indicates that CP7 is likely to represent people who work or study at the Institute.

Lastly, CP8 is the cluster that contains trajectories with the longest time in the Residential area (0.5 to 11 hours in the day and 12 hours at night). Its corresponding transition probability diagram has visually stronger links are between Residential buildings instead of the Facility buildings, as compared to CP1 to CP7, which have very few links to the Residential buildings. There are also a few links between Residential and Facility area, but these are considerably much less than the other clusters as well. The probability of each number of unique locations per trajectory appears to decrease exponentially with the number of places per trajectory increasing from two places up till ten places, as compared to other clusters which appear to be more of a parabola shape. One thing to note for the start and end time distribution plot is that the probability values at 3:00 AM and 2:00 AM are much higher than those of other clusters. These may reflect the trajectories of residents who stay at the Residential area overnight, past the cutoff time of 3:00 AM when the day changes.

\subsection{Temporal Analysis of By-Person Clustering Results}

As part of a deeper analysis of the results of clustering trajectory data by person, we explored the temporal aspect of the clustered data by extracting the temporal data of each CP, similar to the extraction in Section III. We then manually grouped the vectors based on the 4 types of days, namely working Mondays to Thursdays, Fridays/PH eve, Saturdays, and Sundays/PH. We then plotted the average of each group, as well as the minimum and maximum boundaries, which is shown by the shaded area surrounding each line. Below, we highlight two results of our temporal analysis.

\begin{figure}[ht]
	\centering
    \includegraphics[width=\linewidth]{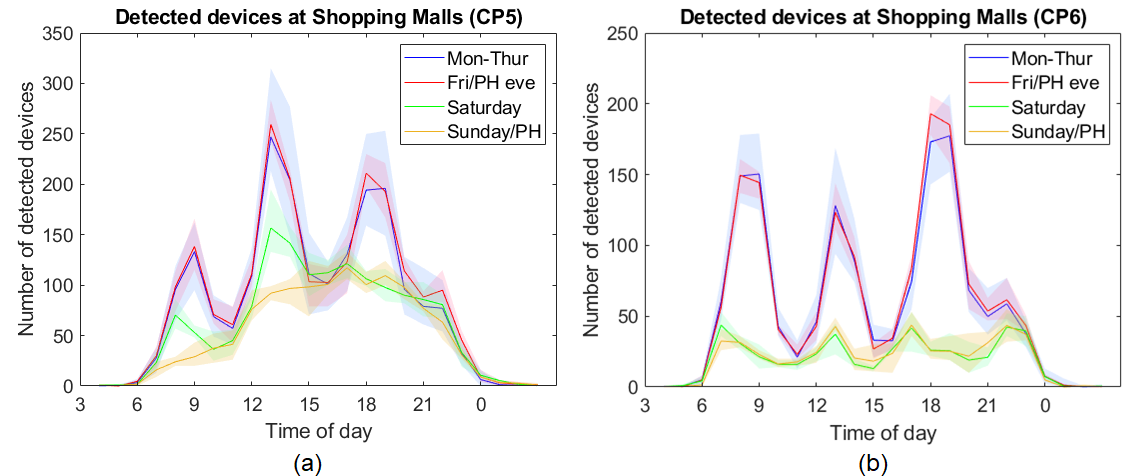}
	\caption{\textit{Daily average number of detected devices from (a) CP5 and (b) CP6 recorded at the Shopping Malls. The shaded area represents the maximum and minimum bounds for each type of day.}}
	\label{fig:ppc-shop}
\end{figure}

Fig.~\ref{fig:ppc-shop} depicts the results from CP5 and CP6, the top two clusters of people recorded as having the longest stay time at the Hospital, as detected at the Shopping Malls. From the proximity of locations, it can be inferred that most of these detections would be of people traveling through the Shopping Malls towards the Hospital, as the extremely long stay time at the Hospital for this cluster's trajectories indicate a high likelihood of the trajectories having the Hospital as their main destination for the day. Upon first glance, it can be seen that for both cases, the Mon-Thur and Friday/PH Eve lines are very close to each other, and much higher than the Saturday and Sunday/PH lines. The Mon-Thur and Friday/PH Eve lines in both cases have three large peaks as well at 7:00-8:00 AM, 12:00 PM, and 5:00-6:00 PM, and one small peak at 9:00 PM. However, the 12:00 PM peak for CP5 is the highest peak, while the 5:00-6:00 PM peak is the highest for CP6.

For CP5, the Saturday line peaks in the morning and at 12, before decreasing gently throughout the rest of the day to nearly zero at 11:00 PM. In contrast, the Sunday/PH line for CP5 lacks distinctive peaks, instead gently curving upwards and then decreasing in a similar way to the Saturday line after 4:00 PM. On the other hand, the Saturday and Sunday/PH lines for CP6 are relatively similar and both have four peaks of approximately equal height spread throughout the day. These peaks occur at 6:00 AM, 12:00 PM, 4:00 PM, and 9:00 PM. The difference in the weekend behavior for these CPs, in addition to the length of time spent at the Hospital for each cluster, supports the line of thinking that CP5 is more likely to represent visitors to the Hospital who go shopping afterwards, while CP6 is likely to represents the people who are employed at the Hospital. 

\begin{figure}[ht]
	\centering
	\includegraphics[width=\linewidth]{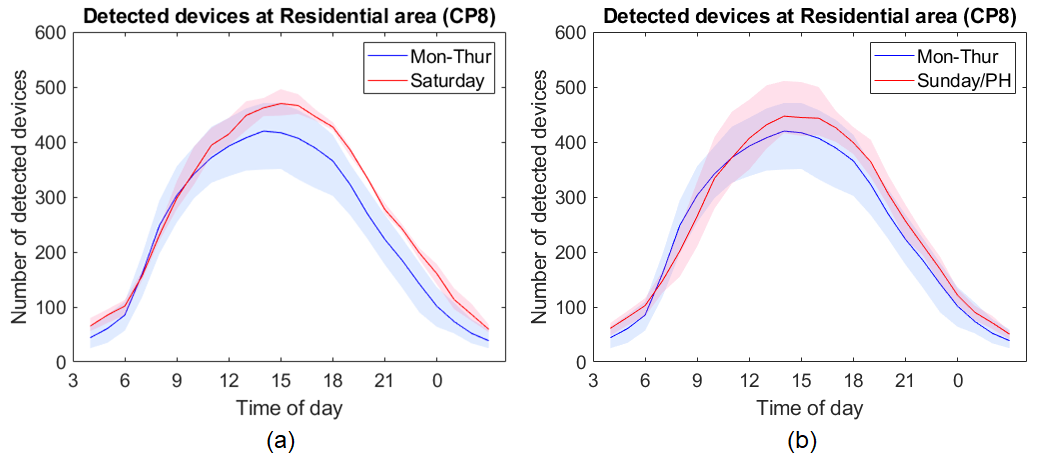}
	\caption{\textit{Daily average number of detected devices from CP8 recorded at the Residential area. Selected lines show comparisons between (a) Mon-Thur and Saturday, (b) Mon-Thur and Sunday/PH. The shaded area represents the maximum and minimum bounds for each type of day.}}
	\label{fig:ppc-hdb}
\end{figure}

Fig.~\ref{fig:ppc-hdb} depicts the results from CP8, the cluster of people recorded as having the longest stay time at the Residential area, as detected in the Residential area. Due to the length of stay at the Residential area, it is likely that they are residents of that area or have that area as a main destination for the day. Since the numbers of detected devices are very similar overall for the four different types of days, two types of days were chosen in each subfigure to do a clear comparison. Fig. 7(a) shows the comparison between working Mondays to Thursdays and Saturdays, while Fig. 7(b) shows the comparison between working Mondays to Thursdays and Sundays/PH.

For Fig. 7(a), it can be seen that more people are detected in the evenings on Saturdays as compared to working Mondays to Thursdays. This can reflect that more people from the Residential area stay out late on Saturday nights as compared to working weeknights. This makes sense if people tend to go home early when there is work the next day, as compared to Saturdays when there is a much lower probability of people going to work on Sundays/PH.

Fig. 7(b) shows that there are more devices detected before 10:00 AM on working Mondays to Thursdays as compared to on Sundays/PH, but there are more devices detected after 10:00 AM on Sundays/PH than on working Mondays to Thursdays. This could indicate that the residents generally wake up or leave the house later in the mornings on Sundays/PH than on working Mondays to Thursdays.

\section{Clustering By Location}

This section addresses the final aspect of clustering addressed in this study, which is clustering by location. The first part of this section presents the results of clustering of the transition probability matrices illustrating transition probabilities between pairs of buildings, while the second part provides a further examination into the transition patterns extracted from the data.

\subsection{Clustering Results}
After clustering the data by temporal patterns as well as by individual patterns, the third aspect of trajectory data clustering is to look at the spatial patterns. The number of detected transitions between each pair of nodes is extracted and compiled into a transition probability matrix, where each row denotes the probability of people moving out of the corresponding source node, and each column denotes the probability of people moving towards the corresponding destination node. For this matrix, we considered 20 locations - 6 from the Facility area as well as 14 individual buildings from the Residential area. The transition probability for each entry of the matrix were calculated using the below equation: 
\begin{equation}
    T(i,j) = \frac{N(i,j)}{\sum_{k\in[1,20],k \neq i} N(i,k) }
\end{equation}

\begin{figure*}[p]
	\centering
	\includegraphics[width=\linewidth]{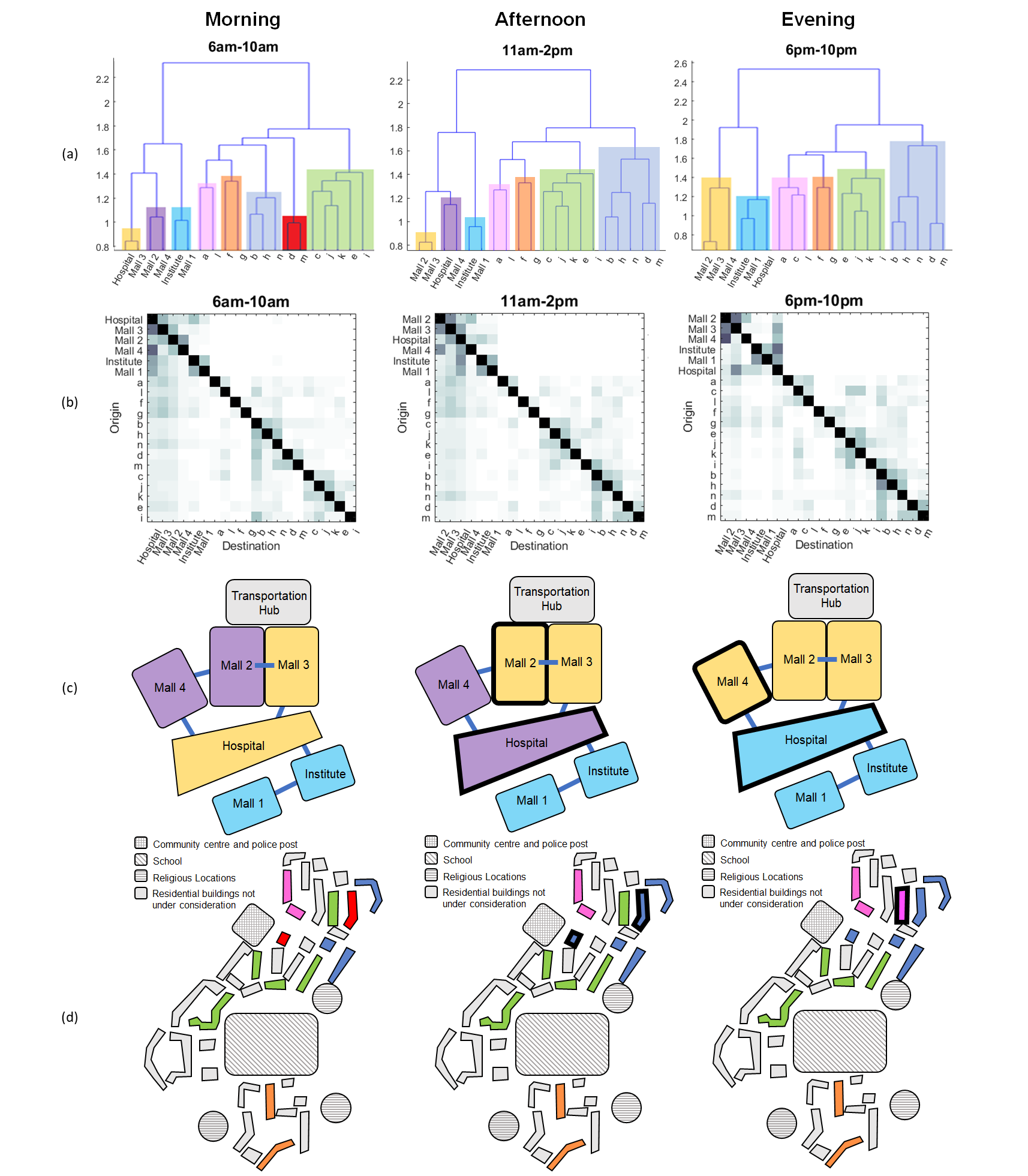}
	\caption{\textit{Diagrams showing different clustering arrangements of locations over the course of the day. (a) Dendrograms produced from HAC, using the transition probability  matrices in (b) as input. (c) Corresponding map of locations in the Facility area. (d) Corresponding map of locations in Residential area. Buildings in grey do not have sensors installed. Buildings that changed from a different grouping in the previous time period have bolded outlines, e.g. afternoon different from morning.}}
	\label{fig:table1}
\end{figure*}

\noindent where $T(i,j)$ refers to the entry of the transition probability matrix in row $i$ and column $j$ for $i \ne j$, and $N(i,j)$ refers to the number of transitions observed in the data moving from node $i$ to node $j$ for $i \ne j$. The diagonal entries of the matrix, indicating probability of each node going back to itself, were then set to 1 to fill up the matrix. As the entries in the input matrix are distances rather than coordinates in a feature space, the use of $k$-means clustering is less suitable. Thus, in this case, the transition probability matrix was then subject to HAC as described in \cite{mullner2011modern}.

HAC is an algorithm designed to cluster data points together based on a given distance matrix. Many different types of linkages can be used such as average linkage, single linkage, Ward linkage, and so on. The result of HAC can be visualized in the form of a dendrogram, which has all the nodes at the bottom as separate 'leaves', which are joined together pairwise by 'branches', until all the clusters are joined together at the very top. The rows and columns of the transition probability matrix would then be rearranged simultaneously according to the output dendrogram. In this case, Ward's method\cite{ward1963hierarchical} is used to calculate the linkages between different clusters and data points at each level. Ward's method is an objective function approach involving the pairing of clusters at each step that results in the minimum increase in the total within-cluster variance after merging. The total within-cluster variance (WCV) is shown in Eqn.~\ref{eq:6}, where $x_i$ represents data point $i$, $j$ represents the cluster number, and $\mu_j$ represents the mean of all the points within cluster $j$. The initial cluster distances are defined as the squared Euclidean distance between points. We hypothesize that the buildings will form roughly spherical clusters based on the map, and Ward's linkage is suitable for this.

\begin{equation}
	WCV = \sum_{j}\sum_{i:ind_i=j}||x_i - \mu_j||^2
	\label{eq:6}
\end{equation}

The above process was done separately for three different time periods of a day for the whole dataset. The first selected time period was between 6:00 AM to 10:00 AM, which is the time when people generally have breakfast, leave their houses, or arrive at their workplace. The second selected time period was between 11:00 AM to 2:00 PM, which is the time when people generally have lunch, and thus there could be a more prominent movement around the malls. Lastly, the third selected time period was between 6:00 PM and 10:00 PM, which is generally the time when people working office hours leave work, have dinner, or return home. The results of HAC by location for each time period are shown in Fig.~\ref{fig:table1}. From a general overview, it can be easily seen that the clustering results differ for each time of the day. 

In Fig.~8(a), the dendrograms show that the groupings of similar buildings change with different timings of the day. The first six labels of each dendrogram represent the Malls, Institute and Hospital. They are grouped into pairs in the morning and afternoon, while they are grouped in threes in the evening. The yellow and purple pairs also change members between morning and afternoon. For a better illustration of the groupings with respect to the building map, a simplified version is shown in Fig.~8(c). The remaining 14 labels of each dendrogram represent the buildings in the Residential area. These also differ throughout the day, and an illustration can be seen in Fig.~8(d). The buildings that were from a different grouping in the previous time period have bolded outlines. The differences between the groupings of the Residential area are described in the following paragraph. 

There is one cluster in the Residential area that stays the same throughout all three time periods, represented by the orange cluster. The buildings that this cluster corresponds to are located on one end of the Residential area and thus they may have similar transition probabilities that differ from the rest of the Residential area. The pink cluster stays the same through the morning and afternoon, but has an additional member in the evening that was originally part of the green cluster. The red cluster present in the morning was grouped with the dark blue cluster for the rest of the day. 

\begin{figure*}[!ht]
	\centering
	\includegraphics[width=\linewidth]{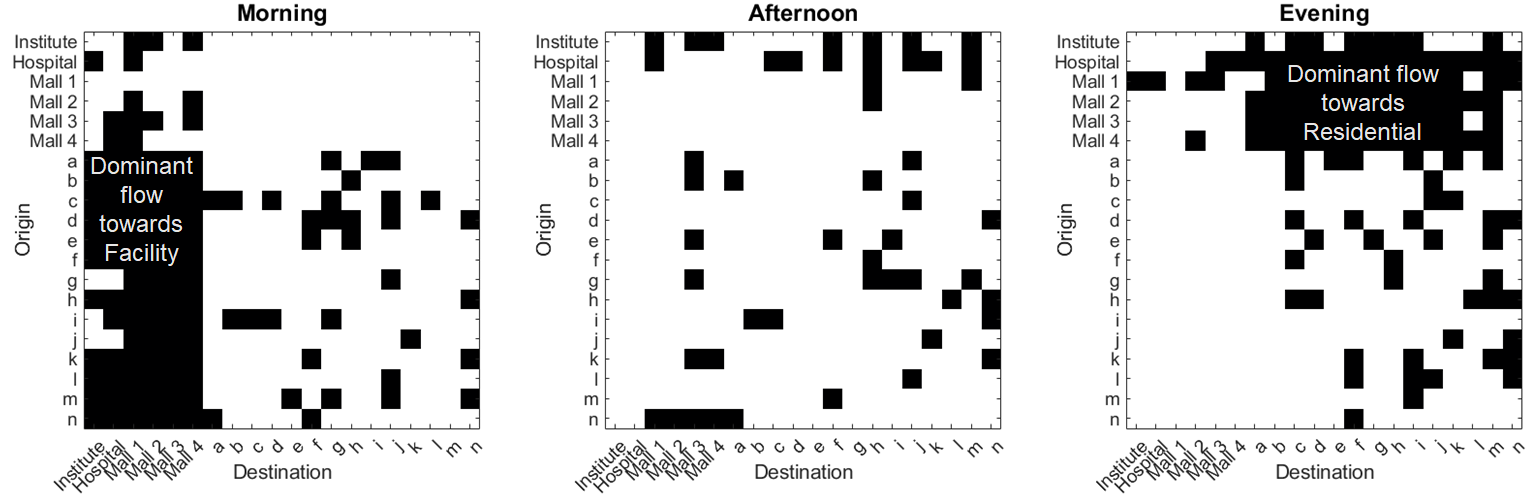}
	\caption{\textit{A plot of dominant directions with probability larger than 0.55. Dominant directions are shown in black.}}
	\label{fig:asymmetry}
\end{figure*}

In Fig.~8(b), the transition probability matrices offer more insight on the general flow of people around the area. One prominent observation is that although there is a visible probability that people from the Residential area move towards the Facility area, there is a very low probability of them moving in the opposite direction. A possible reason for this is that the people coming from the Residential area only contribute to a small percentage of the overall number of visitors to the Facility area, and as a result there is a much larger number of transitions between the Facility buildings as compared to the number of transitions moving from the Facility buildings toward the Residential buildings. This is largely consistent throughout the three time periods. As this large discrepancy makes it difficult to directly identify dominant directions from Fig.~8(b), a clearer visualization of dominant directions will be provided later in Fig.~\ref{fig:asymmetry}.

Another observation is that the direction of the largest transition probabilities, represented by the darkest squares, change over time in the Facility area. In the mornings, the main direction is from Malls 3 and 4 toward the Hospital, as well as from Mall 2 to Mall 4. For the afternoon, the transition probabilities between Malls 2 and 3 are observably higher than those in the morning, and they are higher still in the evening. There is also a larger transition probability in both directions between the Hospital and Mall 3, as well as a higher probability of transition from the Institute to the Hospital. For the evening, two of the main directions are reversed from the morning, coming from Mall 4 to Mall 2, and from Hospital to Mall 3. This could mean that there is an outflow from the transportation hub to the rest of the Facility area in the morning, while the flow is opposite in the evening. This may in turn indicate that the bulk of these detected transitions come from people who work within the Facility area during office hours.

As for the Residential area, it can be seen from Fig.~8(b) that there is a noticeably high probability of arriving at building b from several other buildings since the column of the transition probability matrices corresponding to building b has mostly darker squares than the rest. This trend shows minimal change throughout the time periods. However, there is a lower probability of transitions from the Residential buildings to the Facility buildings in the evening period as compared to the morning and afternoon periods. 

\subsection{Analysis of Transition Patterns}

In order to have a deeper analysis of the spatial patterns, the dominant transition directions between each pair of buildings were plotted in Fig.~\ref{fig:asymmetry}. A direction of transition between a pair of buildings is considered dominant if it occurs with a probability higher than 0.55. This probability is calculated by taking the number of transitions in one direction and dividing by the sum of the number of transitions in both directions. 
At first glance of Fig.~\ref{fig:asymmetry}, one observation that stands out is that in the morning, a large block of the dominant directions start from the Residential area and lead to the Facility area, while it is the reverse case in the evening. This agrees with a general understanding that humans will go out to work in the morning and return home in the evening.

\begin{figure}[!ht]
	\centering
	\includegraphics[width=\linewidth]{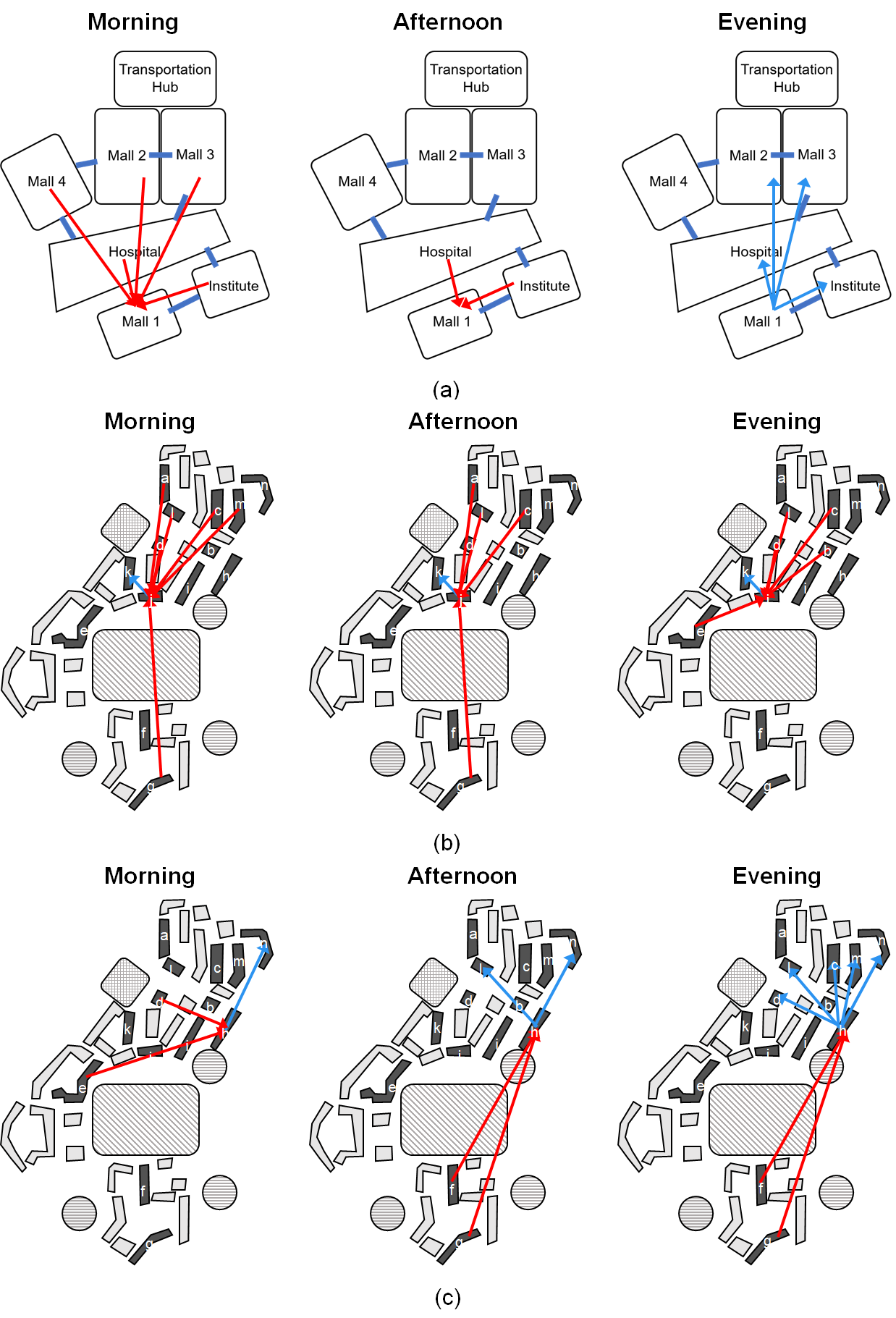}
	\caption{\textit{Highlighted spatial results of analysis of dominant flows over different time periods. (a) Reversal of flows. (b) Continuous outward flow to one specific building. (c) Continuous flow in a general direction. }}
	\label{fig:asymmap}
\end{figure}

The rest of the buildings' flows are analyzed through plotting the dominant directions on maps, focusing on the inflow and outflow of one building at a time, called the building of interest. Patterns can be more easily identified visually, and three such patterns are highlighted in Fig.~\ref{fig:asymmap}. These include reversal of flow at different times of day, continuous flow towards a specific building, and continuous flow in a general direction. Firstly, in Fig.~10(a), the building of interest is Mall 1, located in the Facility area. In the morning, the dominant flows are all inwards towards Mall 1, as shown by the red arrows. Subsequently, the number of dominant flows towards Mall 1 decreases in the afternoon, and finally, in the evening, all the dominant directions are flowing outwards from Mall 1, as shown by the blue arrows.

Next, Fig.~10(b) shows an example of continuous flow towards a specific building throughout the day. The building of interest in this figure is Building $j$, in the Residential area. At all three time periods, the dominant outward flow (indicated by the blue arrow) is always towards its neighboring building, Building $k$, in the Residential area. 

Lastly, Fig.~10(c) shows an example of continuous flow in a general direction throughout the day. The building of interest is Building $h$, another building in the Residential area. Its inflows, shown by red arrows, tend to come from the left and lower parts on the map, while the outflows, shown by blue arrows, tend to go towards the top and right. This pattern appears in all different time periods of the day as well.

\section{Conclusion}

In this paper, we proposed a systematic approach to analyze trajectory data obtained through passive Wi-Fi sensing. We used two unsupervised machine learning techniques, $k$-means clustering and HAC, to examine three different aspects of clustering of trajectory data, namely by time, by person, and by location. In doing so, we observed patterns of daily movement such as the fluctuation of people count over the course of the day, clusters of trajectories belonging to different types of people, as well as the relative volumes of flow between different buildings. We also provided various ways of visualization for the clustering results.

For future work, the proposed approach can be performed on datasets gathered from different locations, such as for different residential estates and facilities, and a comparison can be done to identify differences such as the difference in trajectory patterns of people living in mature estates as compared to newer estates, or estates with different proximities to different sets of facilities. The findings could help in future work related to urban planning in the following ways. Clustering by time gives insights on the daily footfall in various buildings on each type of day, therefore event planning could be more informed, or building tenant proportions could be adjusted if deemed to have an effect on the footfall. Clustering by person gives an idea of rough proportions of travel patterns of users living in an estate, and facilities within the estate can also be adjusted based on need. Finally, clustering by location can help in the planning of the land use in upcoming estates, depending on how the users flow from building to building.


%

\section*{Acknowledgment}

This research is supported by the Singapore Ministry of National Development and the National Research Foundation, Prime Minister's Office under the Land and Liveability National Innovation Challenge (L2 NIC) Research Programme (L2 NIC Award No. L2NICTDF1-2017-4).

Any opinion, findings, and conclusions or recommendations expressed in this material are those of the author(s) and do not reflect the views of the Singapore Ministry of National Development and National Research Foundation, Prime Minister's Office, Singapore.





\bibliographystyle{IEEEtran}
\bibliography{bibliography}
%

%
\begin{IEEEbiography}[{\includegraphics[width=1in,height=1.25in,clip,keepaspectratio]{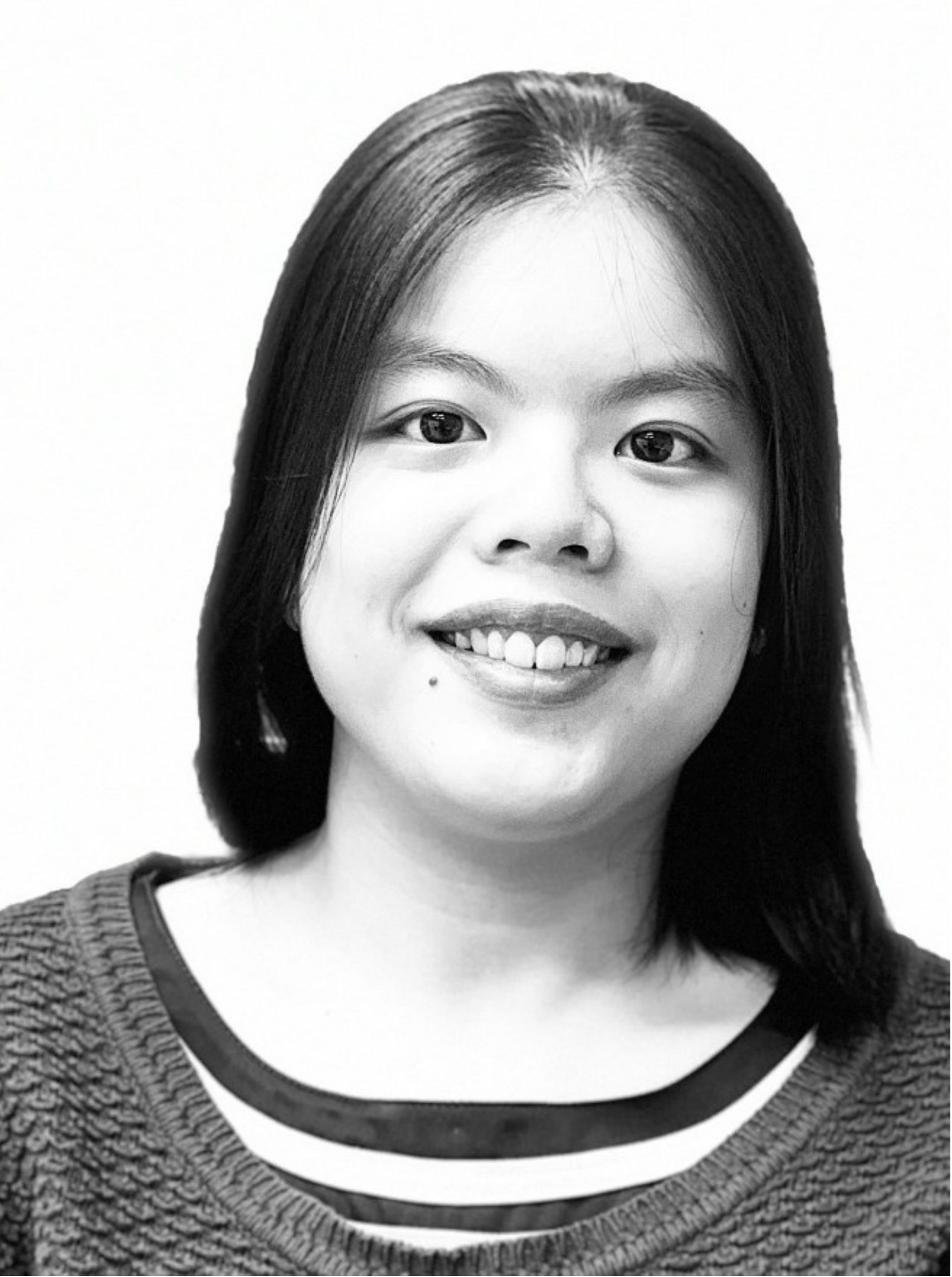}}]{Zann Koh} received the B.Eng degree in Engineering and Product Development from the Singapore University of Technology and Design, Singapore, in 2017. She is currently pursuing the Ph.D. degree with the Singapore University of Technology and Design, Singapore, under Dr. Chau Yuen’s supervision. Her current research interests include big data analysis, data discovery, urban human mobility, and unsupervised machine learning.
\end{IEEEbiography}
\begin{IEEEbiography}[{\includegraphics[width=1in,height=1.25in,clip,keepaspectratio]{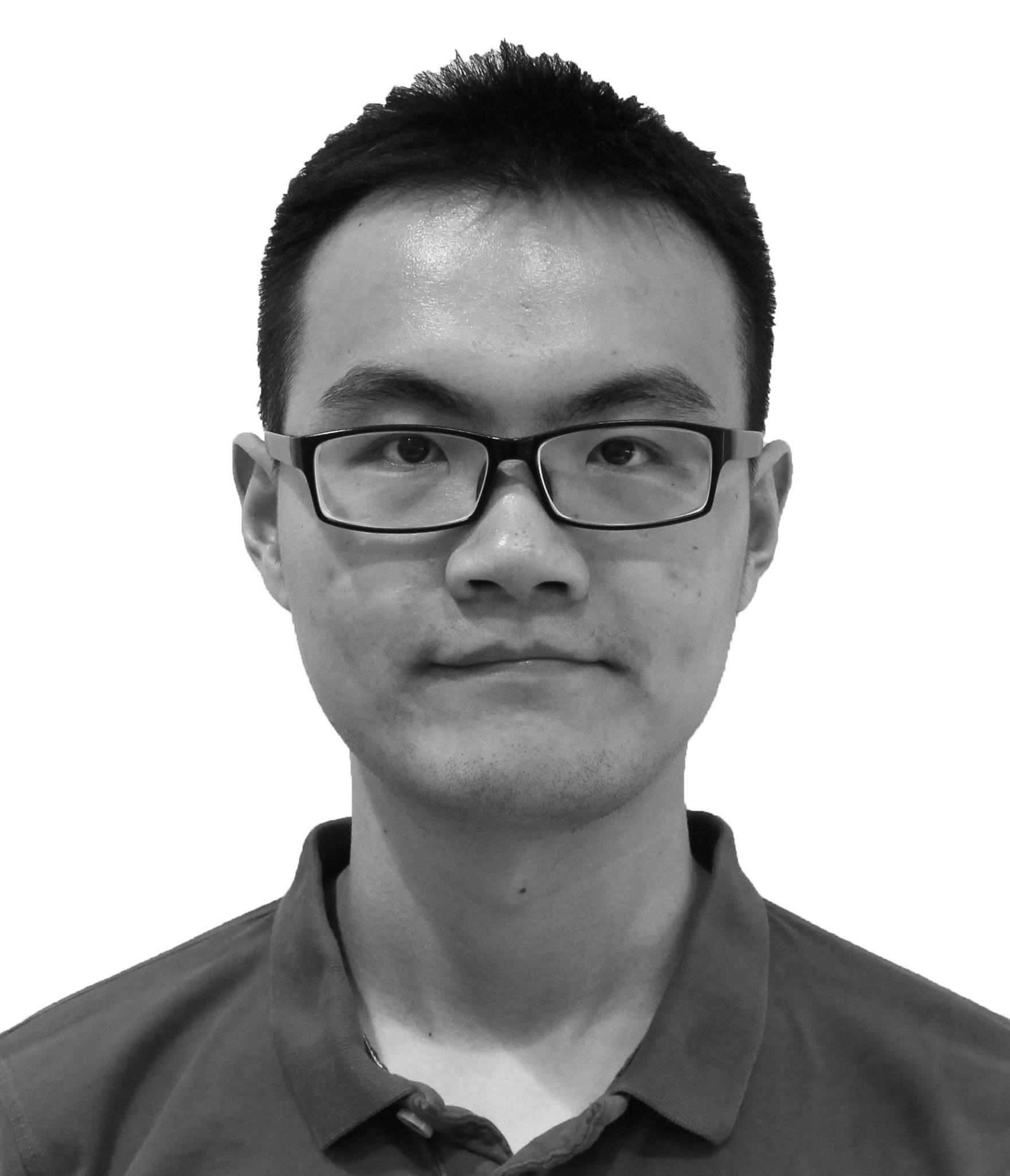}}]{Yuren Zhou} received the B.Eng. degree in Electrical Engineering from Harbin Institute of Technology,	Harbin, China in 2014, and the Ph.D. degree from Singapore University of Technology and Design, Singapore in 2019, with a focus on data mining and smart city applications. He is currently a postdoctoral research fellow at Singapore University of Technology and Design. His current research interests include big data analytics and its application in urban human mobility, building energy management, and Internet of Things.
\end{IEEEbiography}
\begin{IEEEbiography}[{\includegraphics[width=1in,height=1.25in,clip,keepaspectratio]{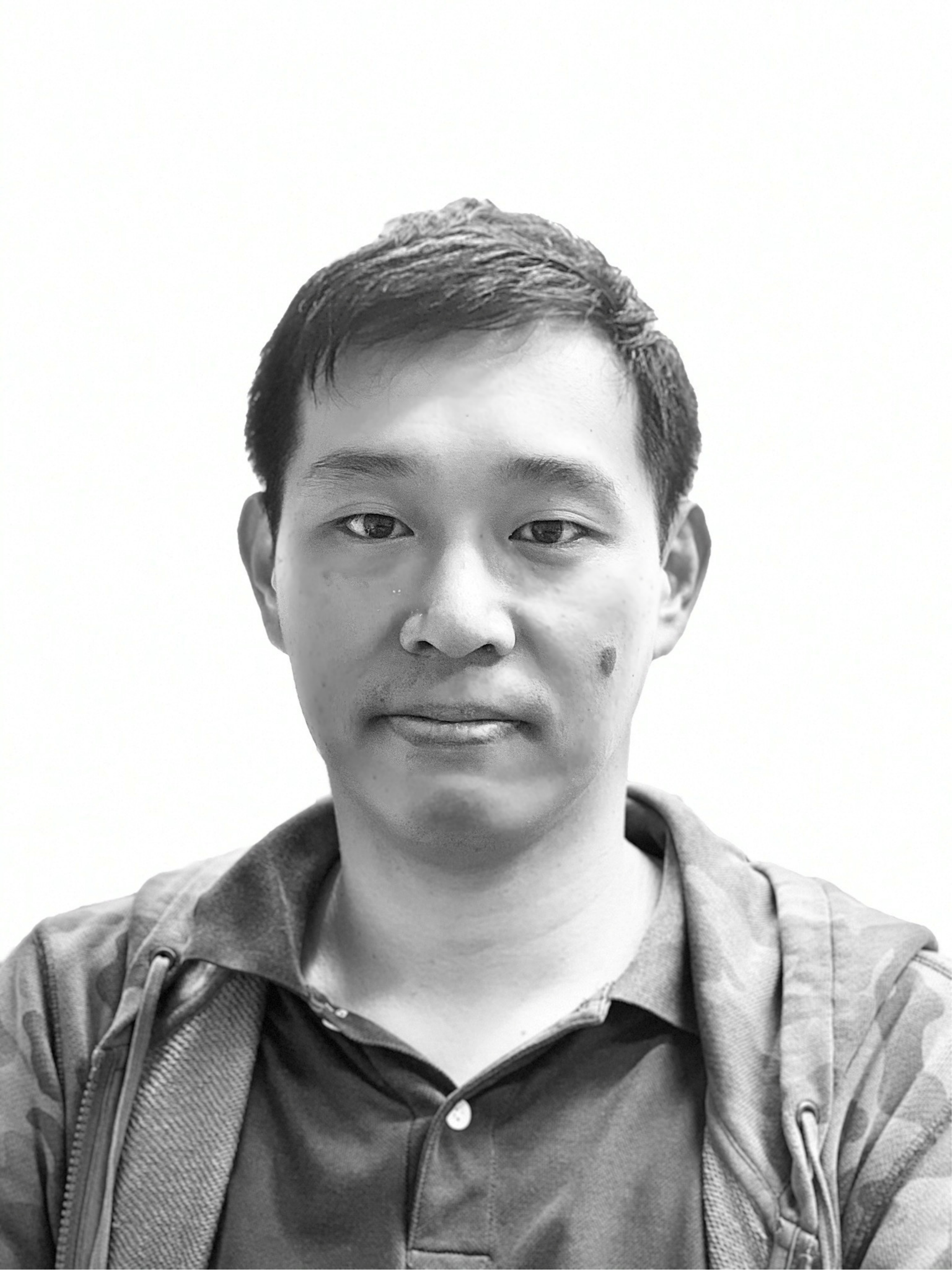}}]{Billy Pik Lik Lau} received the degree in computer science and M.Phil. degree in computer science from Curtin University, Perth, WA, Australia, in 2010 and 2014, respectively. He is currently a Ph.D. candidate with Dr. Chau Yuen at the Singapore University of Technology and Design, Singapore. He studied the cooperation rate between agents in multiagents systems during master studies. His current research interests include urban science, big data analysis, data knowledge discovery, Internet of Things, and unsupervised machine learning.
\end{IEEEbiography}
\begin{IEEEbiography}[{\includegraphics[width=1in,height=1.25in,clip,keepaspectratio]{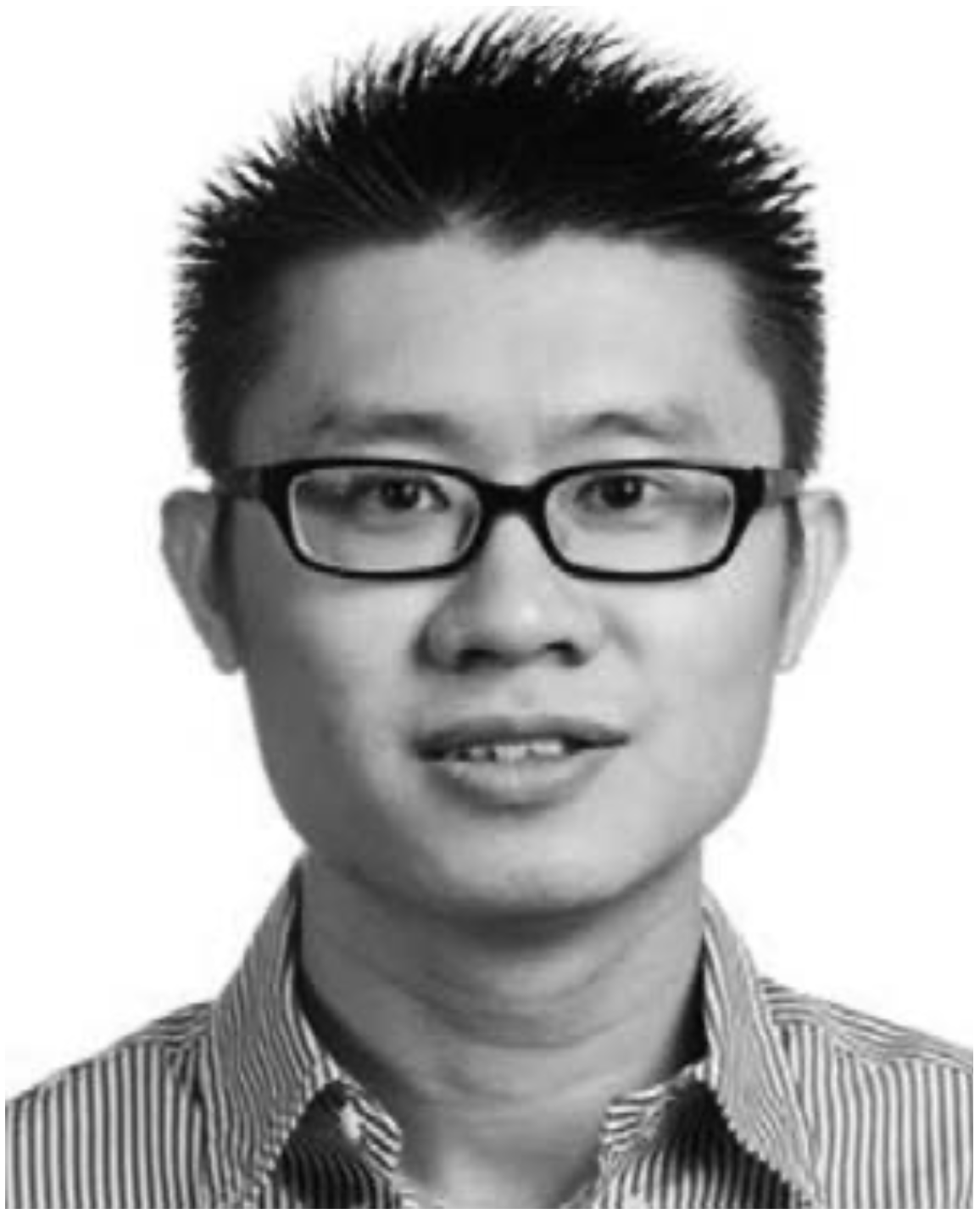}}]{Chau Yuen} is currently an Associate Professor at Singapore University of Technology and Design. He received the B.Eng. and Ph.D. degrees from Nanyang Technological University, Singapore, in 2000 and 2004, respectively. He was a Postdoctoral Fellow at Lucent Technologies Bell Labs, Murray Hill, NJ, USA, in 2005. He was a Visiting Assistant Professor at The Hong Kong Polytechnic University in 2008. From 2006 to 2010, he was a Senior Research Engineer at the Institute for Infocomm Research (I2R, Singapore), where he was involved in an industrial project on developing an 802.11n Wireless LAN system, and participated actively in 3Gpp Long Term Evolution (LTE) and LTE-Advanced (LTE-A) Standardization. He has been with the Singapore University of Technology and Design since 2010. He is a recipient of the Lee Kuan Yew Gold Medal, the Institution of Electrical Engineers Book Prize, the Institute of Engineering of Singapore Gold Medal, the Merck Sharp and Dohme Gold Medal, and twice the recipient of the Hewlett Packard Prize. He received the IEEE Asia-Pacific Outstanding Young Researcher Award in 2012. He serves as an Editor for the IEEE Transaction on Communications and the IEEE Transactions on Vehicular Technology and was awarded the Top Associate Editor from 2009 to 2015.
\end{IEEEbiography}
\begin{IEEEbiography}[{\includegraphics[width=1in,height=1.25in,clip,keepaspectratio]{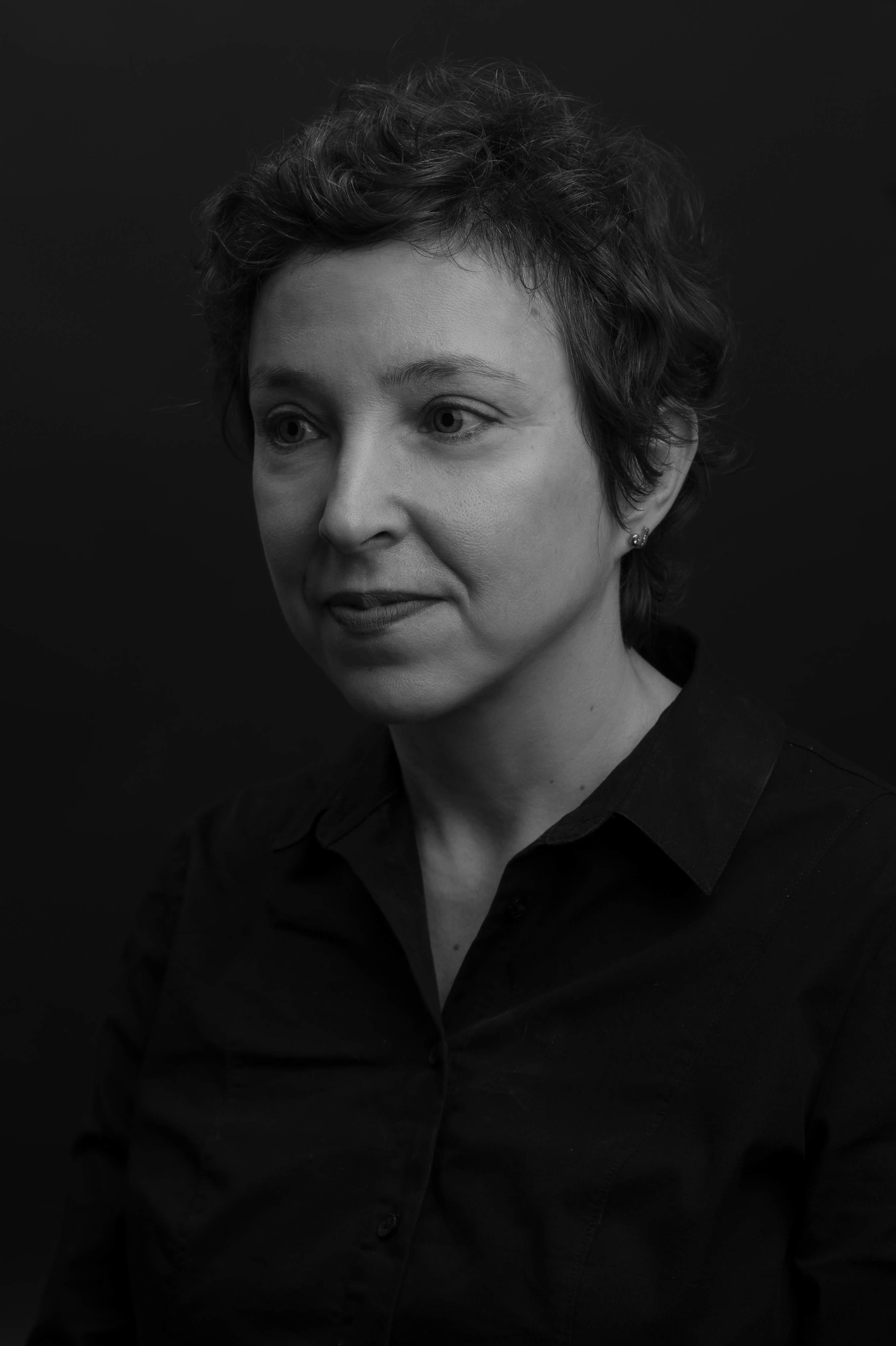}}]{Bige Tun\c cer} is an associate professor at the Architecture and Sustainable Design Pillar of Singapore University of Technology and Design (SUTD), where she founded the Informed Design Lab. The lab’s research focuses on data collection, information and knowledge modeling and visualization, for informed architectural and urban design.
She received her PhD in Architecture from Delft University of Technology (TU Delft), her MSc (computational design) from Carnegie Mellon University, and her BArch from Middle East Technical University.  She was an assistant professor at TU Delft, a visiting professor at ETH Zurich, a visiting scholar at MIT, and a visiting professor at Computer Engineering Department of University of Pavia, Italy. Her research interests include evidence based design, big data informed urban design, and design thinking. She leads and participates in various large multi-disciplinary research projects in evidence informed design, IoT, and big data.
\end{IEEEbiography}
\begin{IEEEbiography}[{\includegraphics[width=1in,height=1.25in,clip,keepaspectratio]{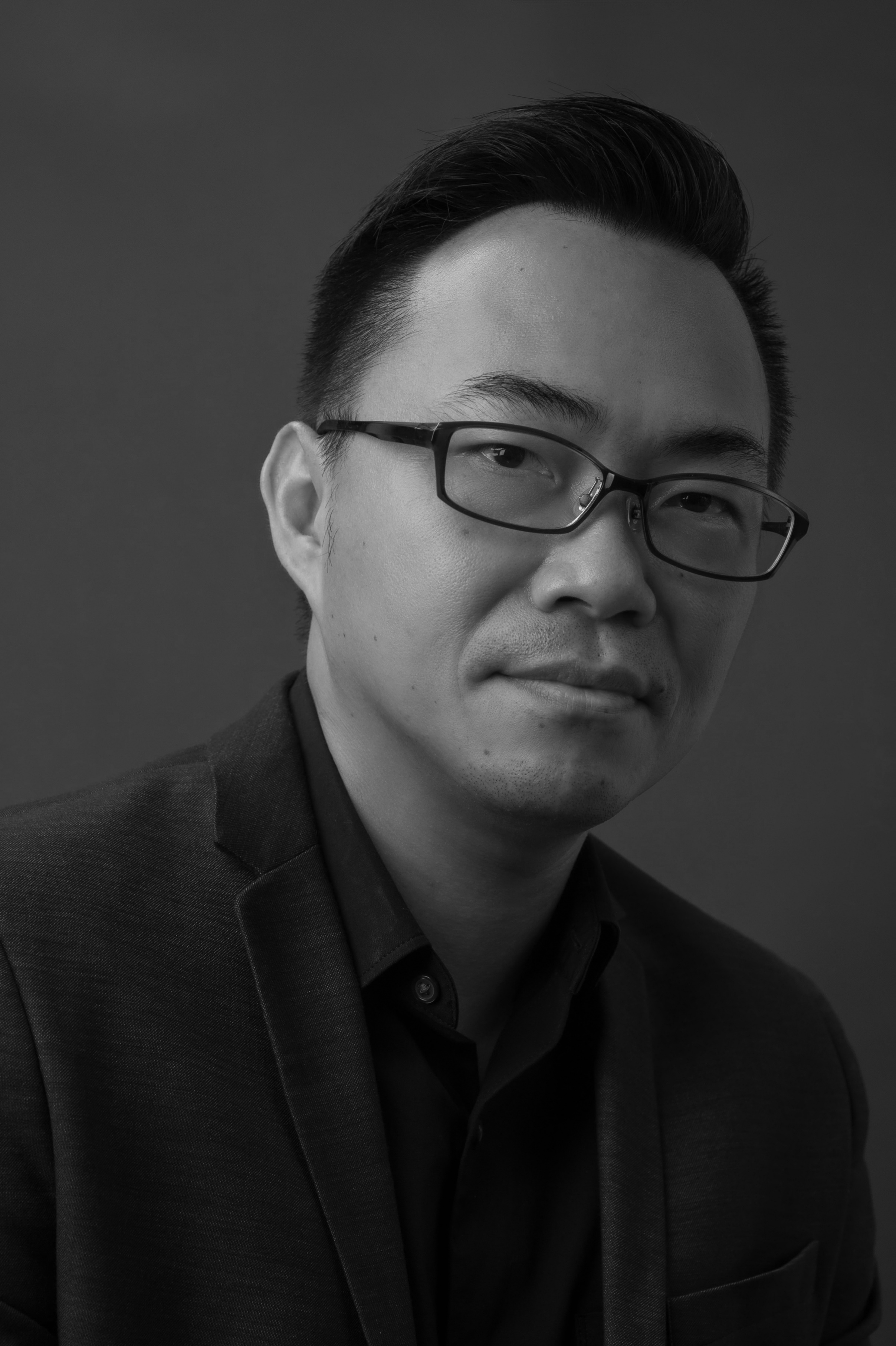}}]{Keng Hua Chong} is Associate Professor of Architecture and Sustainable Design at the Singapore University of Technology and Design (SUTD), where he directs the Social Urban Research Groupe (SURGe) and co-leads the Opportunity Lab (O-Lab). His research on social architecture particularly in the areas of ageing population, liveable place and data-driven collaborative design has led to several key publications and projects, including Creative Ageing Cities, Second Beginnings, and the New Urban Kampung Research Programme.
\end{IEEEbiography}
\vfill






\end{document}